\documentclass{deepwise}

\newcommand{\ReportAuthors}{%
  Ping Gong\textsuperscript{1,*}, Shiyuan Su\textsuperscript{1,*},
  Fandong Zhang\textsuperscript{1}, Xinchen Han\textsuperscript{1},\\
  Haowei Sun\textsuperscript{1}, Yiming Li\textsuperscript{1}, Yizhou Yu\textsuperscript{1,2}%
}
\newcommand{\ReportAffiliations}{%
  \textsuperscript{1}Artificial Intelligence Laboratory, Deepwise Healthcare\\
  \textsuperscript{2}The University of Hong Kong%
}
\newcommand{\ReportAuthorNote}{\textsuperscript{*}Equal contribution.}
\newcommand{\ReportOrganization}{Deepwise Healthcare}
\newcommand{\ReportShortTitle}{SAMI3D-DW Technical Report}
\newcommand{\ReportNumber}{DW-AILAB-TR-2026-001}
\newcommand{\ReportVersion}{1.0}
\newcommand{\ReportDate}{September 23, 2026}
\newcommand{\ReportLogo}{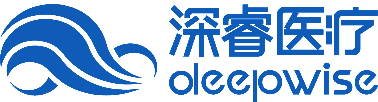}

\title{SAMI3D-DW: Interactive Segmentation of Any 3D Medical Images}
\author{%
  \ReportAuthors\\[0.3em]
  {\small\ReportAffiliations}\\[0.15em]
  {\footnotesize\ReportAuthorNote}%
}
\date{}

\newcommand{\sami}{SAMI3D-DW\xspace}
\newcommand{\dice}{Dice\xspace}

\begin{document}

\maketitle
\begin{abstract}
Interactive segmentation of 3D medical images supports quantitative analysis
of anatomical structures and disease while allowing users to specify and refine their
targets. Despite substantial progress by nnInteractive and VISTA3D, reliable
segmentation across diverse clinical targets remains challenging,
particularly for complex anatomical structures and the heterogeneous, long-tailed spectrum
of pathology.
We present \sami V1 (hereafter \sami), an interactive 3D
segmentation model trained on Deepwise's large-scale proprietary medical
image datasets. We evaluate the model under simulated user interactions on a
CT/MR benchmark comprising 4,326 cases from 219 source datasets, spanning
107 anatomical and pathological categories, organized by a medical taxonomy
and evaluated with a category-balanced DSC score.
\sami achieves the highest category-macro \dice among evaluated methods
in both interaction modes. With one point, it scores 0.5756 versus 0.5316
for nnInteractive, the strongest baseline, rising to 0.7771 versus 0.7495
with five points. With bounding-box initialization, the scores are 0.7129
versus 0.6530. After five corrective clicks, \sami reaches 0.8004 versus
0.7868.
For radiologists and clinicians, \sami enables segmentation of complex
anatomical structures, including intracranial vessel trees on CT and MR angiography,
with a few clicks. In a preliminary in-house comparison involving
neurofibromatosis type 1 (NF1), \sami-assisted tumor annotation took minutes
per case and approximately one-fifteenth of the time required for manual
annotation, highlighting its potential to support volumetric
treatment-response assessment.
\end{abstract}

\begingroup
\setlength{\intextsep}{0pt}
\begin{figure}[!ht]
  \centering
  \includegraphics[width=0.95\linewidth]{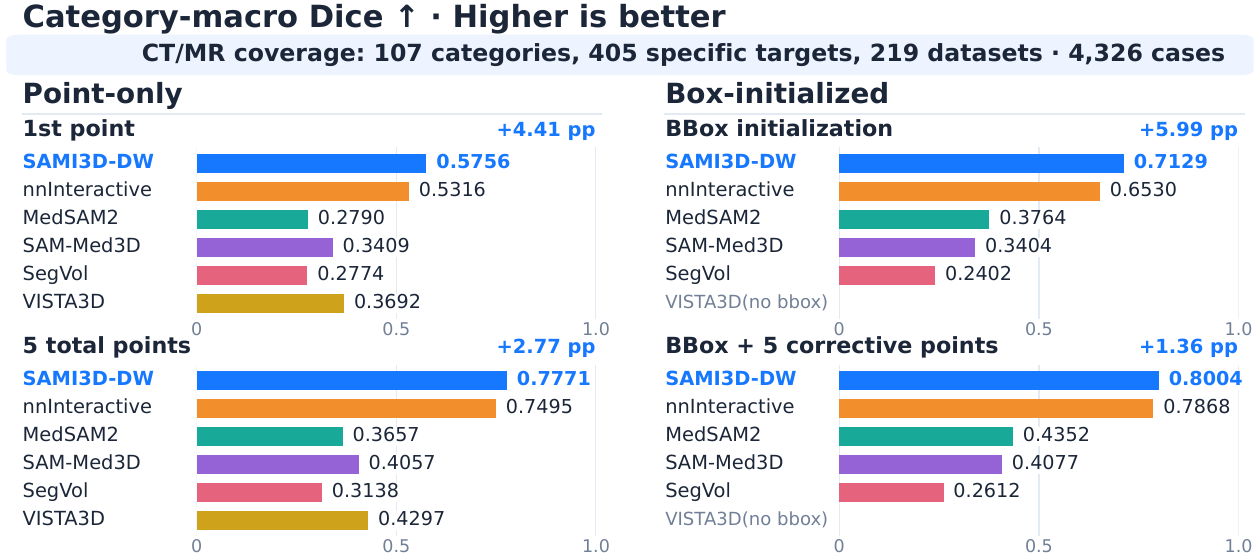}
  \captionsetup{font=footnotesize,skip=3pt}
  \caption{Benchmark-wide performance with point-only and
  box-initialized interaction. Top: initial prompts. Bottom: five total
  points or five corrective points after a BBox. Bar endpoints show
  category-macro \dice on shared 0--1 axes; colors identify
  methods. Gains over SOTA are shown in percentage points (pp)
  at the top-right of each panel.}
  \label{fig:sami3d-overview}
\end{figure}
\endgroup

\clearpage
\tableofcontents
\clearpage

\section{Introduction}
\label{sec:introduction}

Medical image segmentation delineates anatomical structures and pathological
regions for quantitative analysis, treatment planning, and longitudinal
assessment. Task-specific methods such as U-Net and nnU-Net have enabled
accurate segmentation across many applications
\citep{ronneberger2015unet,isensee2021nnunet}, but extending these systems to
new targets or imaging settings generally requires additional annotation,
training, and validation. Interactive segmentation offers a flexible way to
specify the intended target and refine its boundaries at inference time.

SAM and MedSAM established widely used spatial prompting interfaces
\citep{kirillov2023segmentanything,ma2024medsam}. SAM-Med3D and SegVol extended
promptable segmentation to volumetric medical images
\citep{wang2024sammed3d,du2024segvol}.
VISTA3D integrates automatic segmentation with interactive editing, MedSAM2
adapts memory-based propagation to medical volumes and videos, and
nnInteractive supports diverse prompts and iterative correction in 3D
\citep{he2025vista3d,ma2025medsam2,isensee2025nninteractive}.
A central question for general-purpose use is how accurately these models
segment diverse targets, including complex anatomical structures and heterogeneous
pathology, from an initial prompt and how they respond to subsequent
correction.

We present \sami, a model for interactive segmentation of
3D medical images, trained on Deepwise's proprietary medical image datasets.
The model produces volume masks from spatial prompts and supports point
and two-dimensional bbox interaction.

Evaluation must reflect both target diversity and aggregation choices.
A few familiar datasets may leave relevant targets unrepresented, while
dataset-level averages alone do not reveal repeated coverage of the same
structures. Pooling all objects gives more weight to categories with more
examples. A medical taxonomy supplies common reporting categories across
sources, making anatomical and pathological coverage explicit.

We therefore evaluate \sami on a benchmark comprising 4,326 cases from
219 source datasets and 107 medical categories, focusing on computed
tomography (CT) and magnetic resonance (MR), the most common modalities for
clinical volumetric imaging. Dataset labels are harmonized through a medical taxonomy.
Controlled sampling limits contributions from large datasets, while
category-macro \dice weights category means equally.
Both interaction modes use simulated prompts, with
performance reported from initialization through successive corrections
(\cref{sec:evaluation-protocol}).

This report makes two principal contributions:
\begin{itemize}
  \item We introduce an evaluation framework based on a medical taxonomy,
  with broad, fine-grained coverage of anatomical and pathological
  segmentation tasks, and use it to benchmark leading open-source
  interactive segmentation models.
  \item \sami matches or exceeds current state-of-the-art performance on
  this benchmark in both point and bounding-box interaction modes.
  Category-macro \dice reaches 0.5756 with one point and 0.7771 with five
  points, and 0.7129 with box initialization and 0.8004 after five corrective
  points (\cref{sec:main-results}).
\end{itemize}

\clearpage
\section{Related Work}
\label{sec:related-work}

\begin{figure}[p]
  \centering
  \includegraphics[width=\textwidth]{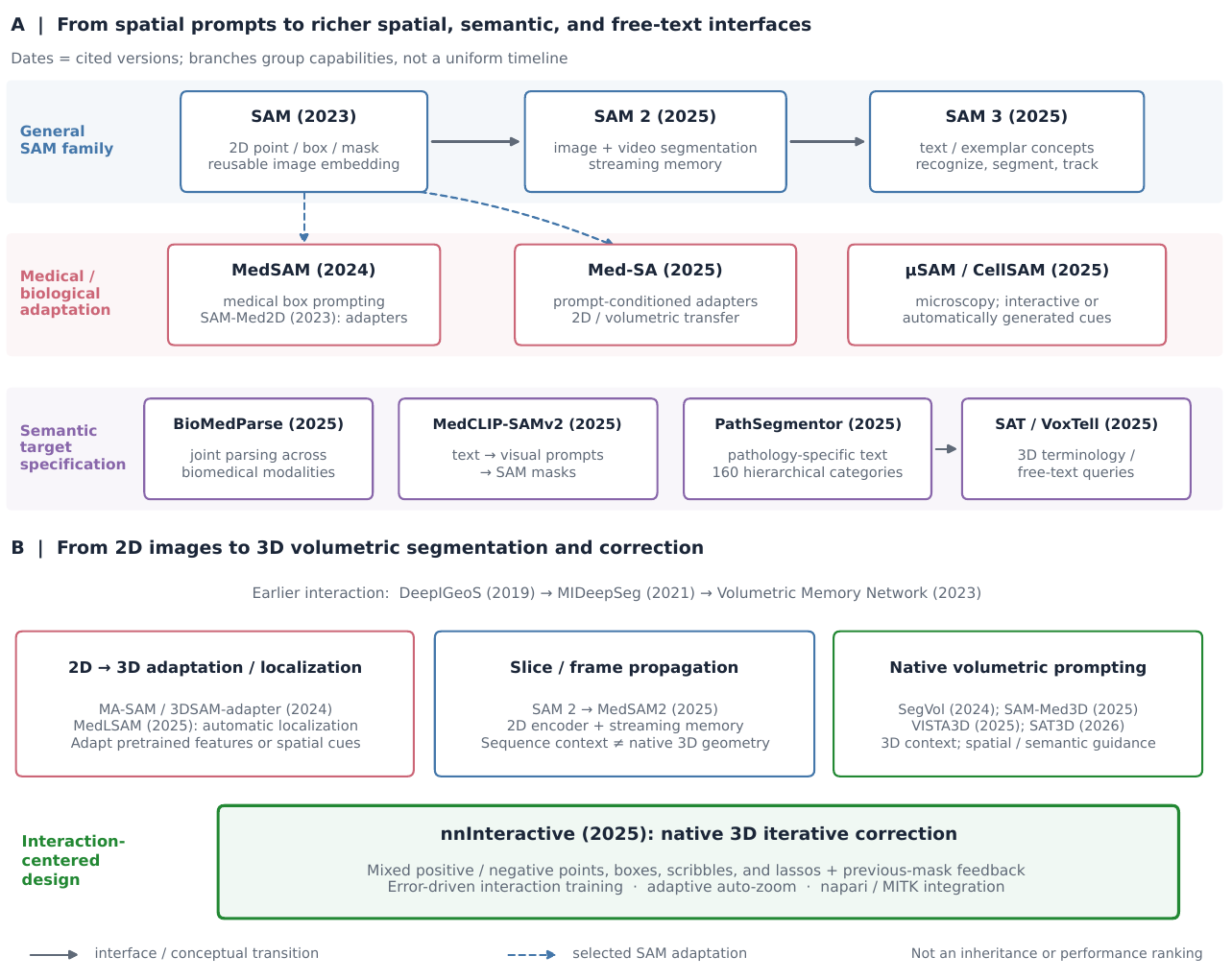}
  \caption{Two major trends in interactive medical segmentation:
  \textbf{A}, from spatial prompts to richer spatial, semantic, and free-text
  interfaces; \textbf{B}, from 2D images to 3D volumetric segmentation and
  correction, the direction pursued by \sami. Arrows indicate conceptual
  transitions or selected adaptations, not performance rankings or direct
  architectural inheritance. Dates follow the cited publication or preprint
  version; grouped branches are not a uniform timeline.}
  \label{fig:interactive-segmentation-evolution}
\end{figure}

\paragraph{From spatial prompts to richer spatial, semantic, and free-text interfaces.}
\Cref{fig:interactive-segmentation-evolution}A summarizes the expansion of
target specification. SAM established a reusable point-, box-, and
mask-prompted interface \citep{kirillov2023segmentanything}. Medical evaluations
exposed domain gaps \citep{mazurowski2023sammedical,huang2024sammedical},
motivating MedSAM's large-scale, box-conditioned medical adaptation and
adapter-based methods \citep{ma2024medsam,cheng2023sammed2d,wu2025medicalsamadapter}.
ScribblePrompt, $\mu$SAM, and CellSAM further developed biomedical interaction
and automatic prompt generation
\citep{wong2024scribbleprompt,archit2025microsam,marks2025cellsam}.
Semantic specification emerged through SAM 3's concept prompts
\citep{carion2025sam3}, MedCLIP-SAMv2's text-to-spatial prompting
\citep{koleilat2025medclipsamv2}, BioMedParse's joint segmentation, detection,
and recognition across biomedical modalities \citep{zhao2025biomedparse},
and PathSegmentor's pathology-specific natural-language interface
\citep{chen2025pathsegmentor}. SAT and VoxTell extend medical terminology and
free-text queries to volumes \citep{zhao2023sat,rokuss2025voxtell}.
These developments broaden how users express \emph{what} to segment, but
semantic intent does not necessarily determine an exact instance or boundary.
Customer feedback at Deepwise and current technical limitations motivate
retaining spatial prompts as a necessary tool for fast, high-accuracy
interactive annotation: points, boxes, and other spatial cues provide direct
localization and precise error correction, even when language supplies the
initial target description.

\paragraph{From 2D images to 3D volumetric segmentation and correction.}
An equally important trend is the transition from slice-wise masks to
volume-level prediction and refinement
(\cref{fig:interactive-segmentation-evolution}B). Earlier geodesic and
memory-based systems already studied correction and unseen-target transfer
\citep{wang2018deepigeos,luo2021mideepseg,zhou2023volumetricmemory}.
MA-SAM, 3DSAM-adapter, and MedLSAM extend pretrained features or localization
to volumes \citep{chen2024masam,gong2024threedsamadapter,lei2025medlsam};
SAM-Med3D, SegVol, VISTA3D, and SAT3D develop native volumetric prompting
\citep{wang2024sammed3d,du2024segvol,he2025vista3d,peiris2026sat3d}.
SAM 2 and MedSAM2 reduce repeated initialization through memory-based
slice propagation \citep{ravi2025sam2,ma2025medsam2}. However, propagation
remains slice-based: changes in anatomical structures and accumulated errors can still
require inspection and correction across many slices. It does not by itself
adequately resolve the multi-slice annotation burden motivating this report.
nnInteractive is the state-of-the-art reference most closely aligned with
this volume-level correction objective \citep{isensee2025nninteractive}.
Its native 3D model turns points, boxes, scribbles, and lassos placed in
2D orthogonal views into full-volume masks, using previous-mask feedback
and error-driven training to support iterative refinement.
\sami continues this second line of development, focusing on
\emph{3D volumetric segmentation and correction driven by 2D spatial prompts}.
We evaluate point-only and box-initialized corrective-point workflows,
measuring full-volume quality over successive interactions rather than
isolated prompted masks, consistent with interactive benchmarking practice
\citep{ulrich2025radioactive}.

\clearpage

\section{Data and Training Overview}
\label{sec:data-training}

\subsection{Training Dataset Curation}
\label{sec:training-data-curation}

\begin{figure}[!htbp]
  \centering
  \includegraphics[width=\textwidth]{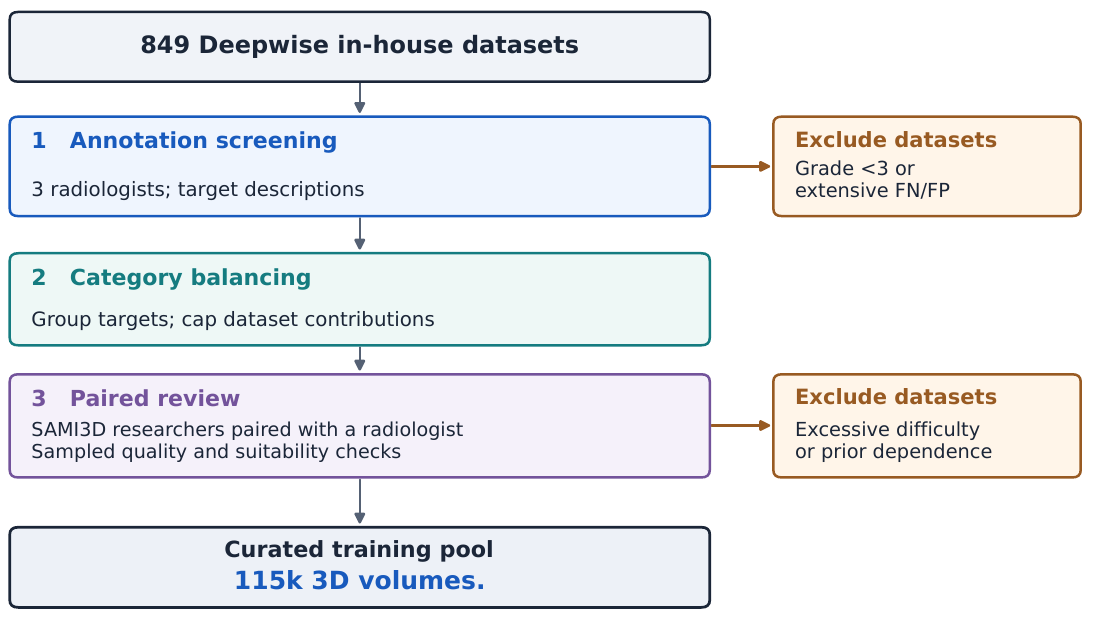}
  \caption{Training data curation. FN and FP denote false-negative and
  false-positive annotation errors.}
  \label{fig:training-data-curation}
\end{figure}

The candidate training collection included 849 Deepwise in-house datasets.
Three-stage curation yielded 115k 3D volumes
(\cref{fig:training-data-curation}).

\textbf{Annotation screening.}
Three radiologists of attending rank or above assessed false-negative (FN)
and false-positive (FP) annotation errors, graded contour quality from 1 to 5
(\cref{tab:training-annotation-rubric}), and described each target. Datasets with
contour grades below 3 or extensive FN/FP errors were excluded.

\textbf{Category balancing.}
Datasets were grouped by target description. Sampling caps limited each
dataset's contribution to balance target categories.

\textbf{Paired review.}
SAMI3D researchers paired with a radiologist reviewed sampled annotations
for quality and training suitability. Datasets judged too
difficult for the training objective (e.g., soft-plaque delineation) or
reliant on expert anatomical conventions (e.g., hepatic segments) were
excluded.

\begin{table}[!htbp]
  \centering
  \caption{Contour-quality grading. Cases graded 1--2 were excluded.}
  \label{tab:training-annotation-rubric}
  \small
  \setlength{\tabcolsep}{5pt}
  \renewcommand{\arraystretch}{1.15}
  \begin{tabularx}{\textwidth}{@{}>{\raggedright\arraybackslash}p{0.20\textwidth}X@{}}
    \toprule
    Grade & Criteria \\
    \midrule
    \textbf{5: Excellent} & Accurate boundaries with fine anatomical detail preserved. \\
    \textbf{4: Good} & Minor focal contour deviations. \\
    \textbf{3: Acceptable} & Overall anatomical structures preserved; localized errors requiring correction. \\
    \textbf{2: Poor} & Major contour errors or substantial anatomical distortion. \\
    \textbf{1: Unusable} & Absent required annotations, invalid masks, or image--mask mismatch. \\
    \bottomrule
  \end{tabularx}
\end{table}

\FloatBarrier

\subsection{Model Training}
\label{sec:model-training}

The architecture and detailed training configuration of \sami are not
publicly disclosed at this stage. The model was trained from scratch on the
Deepwise-curated datasets described above, without using any pretrained
weights.

Training comprised two stages: initial training on the full curated dataset
collection, followed by fine-tuning at a reduced learning rate on a subset
of challenging datasets, particularly those involving lesions.

\section{Evaluation Protocol}
\label{sec:evaluation-protocol}

We evaluate \sami retrospectively under two simulated interaction modes. The
point-only mode uses one to five cumulative foreground/background points and
supports the broadest comparison across model interfaces. The
BBox-initialized mode begins with one two-dimensional (2D) box and permits up
to five corrective point prompts. Its correction sequence follows the
structure of the
\href{https://www.codabench.org/competitions/5263/}{CVPR 2025 Foundation
Models for Interactive 3D Biomedical Image Segmentation Challenge}; the
prompt construction and aggregation rules below define the protocol used in
this report, and the two protocols are not identical. Results from the two
interaction modes are reported separately.

\subsection{Test Dataset}
\label{sec:evaluation-cohorts}

\begin{figure}[htbp]
  \centering
  \includegraphics[width=\textwidth]{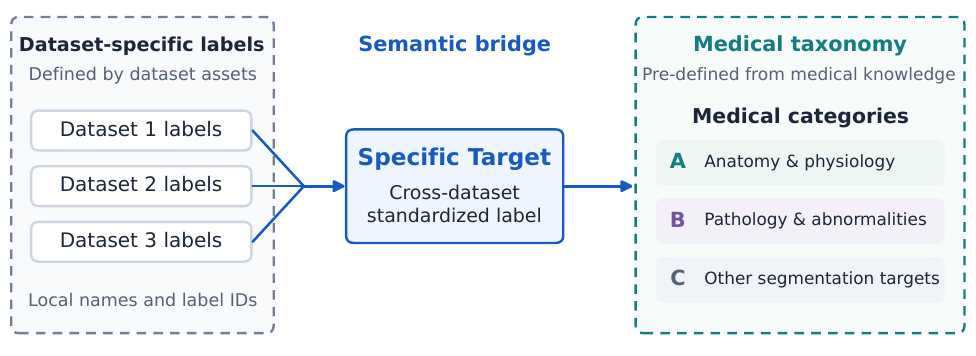}
  \caption{From dataset-specific labels to medical categories.
  Dashed boxes distinguish labels determined by dataset assets (left) from
  categories pre-defined using medical knowledge (right).
  Specific Targets provide shared, standardized definitions across datasets
  and bridge the two spaces. Medical categories are organized under domains
  A (anatomy), B (pathology), and C (other).}
  \label{fig:label-taxonomy-mapping}
\end{figure}

\paragraph{Medical taxonomy.}
Combining medical segmentation datasets requires more than concatenating
their label lists. The same structure may have different names across
sources, whereas similar names may refer to different anatomical structures
or pathological conditions, or reflect differences in laterality or
annotation policy. We distinguish dataset-specific labels
from the pre-defined medical taxonomy (\cref{fig:label-taxonomy-mapping}).
Each dataset defines its own label names and annotation scope.
A \emph{Specific Target} (hereafter target) is a cross-dataset
standardized label that bridges these two spaces. It identifies the medical
object together with its modality, laterality, anatomical or pathological
subdivision, and verified imaging sequence or phase, where available.
Semantically equivalent dataset labels can map to a shared target.
Targets are then grouped into pre-defined reporting categories, organized
by medical domain using medical knowledge rather than dataset identity.
Available annotations determine which parts of the taxonomy are represented,
not how its categories are defined.

\FloatBarrier

The taxonomy comprises three domains. \emph{A: Anatomical and physiological
structures} covers normal human anatomical structures and physiological constituents,
including the liver, kidneys, heart, vascular trees, vertebrae, and muscles.
\emph{B: Pathology and abnormalities} covers disease-related lesions,
abnormal tissues, and abnormal regions, including tumours such as
neurofibromas, haemorrhage, infarction, and infectious lesions.
\emph{C: Other segmentation targets} covers targets outside normal human
anatomical structures and human pathology, including non-human structures,
medical implants, and foreign bodies. Assignment
follows the annotated target: a liver mask belongs to A, whereas a liver
tumour mask belongs to B, even when both originate from the same image.

Within A, structures are grouped by organ system and tissue family. Within B,
lesions are grouped by pathological process, including neoplastic,
infectious, inflammatory and immune-mediated, vascular, traumatic, focal
non-neoplastic, and indeterminate abnormalities. These domain definitions
describe the scope of the taxonomy; benchmark coverage is determined by the
test-set annotations and label-to-target mappings.
\Cref{tab:testset-taxonomy} summarizes the domain definitions,
illustrative examples, and counts for the represented families.

\begin{table}[tbp]
  \centering
  \caption{Medical taxonomy and coverage of the scored test set. Domain
  definitions and illustrative examples describe the classification scope;
  counts refer only to represented families. Instance counts are deduplicated
  within each category; a physical object can contribute to more than one category.}
  \label{tab:testset-taxonomy}
  \small
  \setlength{\tabcolsep}{4.0pt}
  \renewcommand{\arraystretch}{1.05}
  \begin{tabularx}{\textwidth}{@{}l>{\raggedright\arraybackslash}Xrr@{}}
    \toprule
    Code & Taxonomy family & Categories & Instances \\
    \midrule
    \multicolumn{4}{@{}l}{\textbf{A: Anatomical and physiological structures}} \\
    \multicolumn{4}{@{}>{\raggedright\arraybackslash}p{\textwidth}@{}}{%
      \emph{Scope:} Normal human anatomical structures and physiological constituents.
      \emph{Examples:} Liver, kidneys, heart, vascular trees, vertebrae, and muscles.} \\
    \addlinespace[0.2em]
    A01 & Nervous system & 4 & 224 \\
    A02 & Cardiovascular system & 15 & 1,107 \\
    A03 & Respiratory system & 2 & 463 \\
    A04 & Digestive system & 9 & 1,263 \\
    A05 & Urinary system & 4 & 481 \\
    A06 & Reproductive system & 2 & 97 \\
    A07 & Endocrine system & 3 & 296 \\
    A08 & Lymphatic, immune, and hematopoietic systems & 2 & 232 \\
    A09 & Skeletal and articular system & 14 & 2,813 \\
    A10 & Muscular system & 6 & 462 \\
    A12 & Sensory system & 2 & 112 \\
    \cmidrule(l){2-4}
    & \textit{A subtotal} & \textit{63} & \textit{7,550} \\
    \addlinespace[0.35em]
    \multicolumn{4}{@{}l}{\textbf{B: Pathology and abnormalities}} \\
    \multicolumn{4}{@{}>{\raggedright\arraybackslash}p{\textwidth}@{}}{%
      \emph{Scope:} Disease-related lesions, abnormal tissues, and abnormal regions.
      \emph{Examples:} Tumours (including neurofibromas), haemorrhage, infarction,
      and infectious lesions.} \\
    \addlinespace[0.2em]
    B01 & Neoplastic lesions & 23 & 3,579 \\
    B02 & Infectious lesions & 1 & 428 \\
    B03 & Non-infectious inflammatory and immune-mediated lesions & 1 & 149 \\
    B04 & Vascular and circulatory disorders & 2 & 634 \\
    B05 & Trauma and tissue injury & 2 & 238 \\
    B07 & Non-neoplastic focal lesions & 1 & 84 \\
    B10 & Indeterminate imaging abnormalities & 13 & 1,282 \\
    \cmidrule(l){2-4}
    & \textit{B subtotal} & \textit{43} & \textit{6,394} \\
    \addlinespace[0.35em]
    \multicolumn{4}{@{}l}{\textbf{C: Other segmentation targets}} \\
    \multicolumn{4}{@{}>{\raggedright\arraybackslash}p{\textwidth}@{}}{%
      \emph{Scope:} Targets outside normal human anatomical structures and human pathology.
      \emph{Examples:} Non-human structures, medical implants, and foreign bodies.} \\
    \addlinespace[0.2em]
    C99 & Other uncategorized targets & 1 & 25 \\
    \midrule
    \multicolumn{2}{@{}l}{\textbf{Total}} & \textbf{107} & \textbf{13,969} \\
    \bottomrule
  \end{tabularx}
\end{table}

\paragraph{Benchmark coverage and sampling.}
The benchmark is designed to cover diverse anatomical and pathological
targets across body regions in CT and MR. Sampling is controlled
across source datasets to limit dominance by large datasets and commonly used
targets. Medical categories, rather than dataset labels, define the
principal reporting units, and category means receive equal weight. This
combination broadens coverage beyond a small set of familiar datasets and
keeps less frequent target categories visible in the overall evaluation.
Balanced weighting does not imply equal sample counts across categories or
modalities, or a test distribution matching clinical prevalence.

Because this report studies interactive \emph{3D} segmentation, the pool of
candidate test datasets is first filtered at the dataset level
to remove all two-dimensional datasets. We further restrict evaluation to
CT and MR, removing additional 3D datasets from other modalities.

The resulting benchmark comprises 219 source datasets, 4,326 test cases, and
13,826 physical target objects. The curated mapping covers 405 specific targets
and 107 categories with 13,969 associated instances: 63 categories of
anatomical structures with 7,550 associated instances, 43 pathology categories
with 6,394 associated instances, and one Other category with 25 associated
instances. Each physical object is counted once per category, so the
category-level instance total can exceed the physical-object count when an
object maps to multiple medical concepts.

\begin{figure}[tbp]
  \centering
  \includegraphics[width=\textwidth]{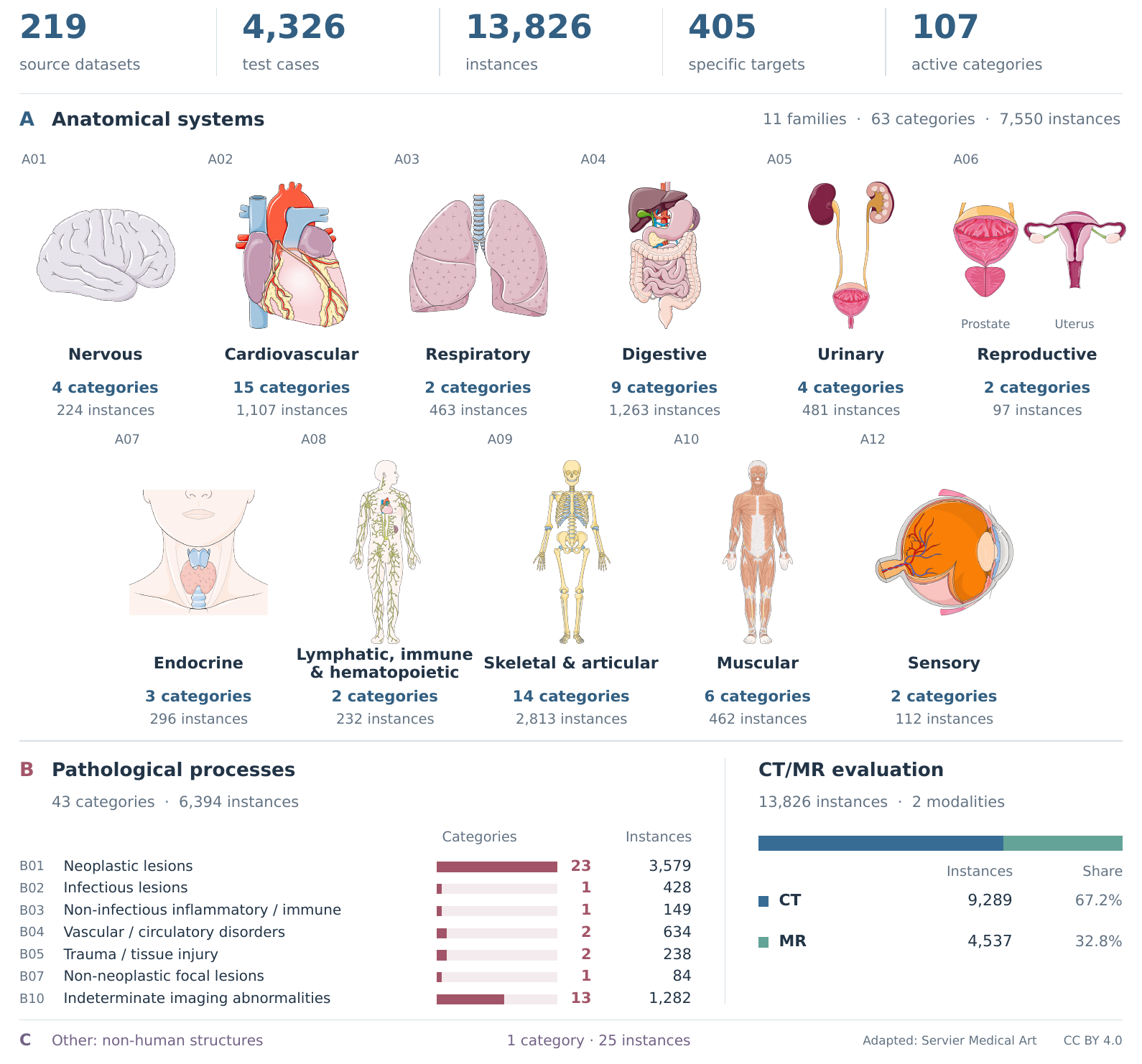}
  \caption{Anatomical, pathological, and modality coverage of the benchmark.
  Medical illustrations show representative anatomical structures for all 11 represented
  anatomical families (A). The pathology panel (B) lists seven families, with
  bar lengths indicating category counts; the Other domain (C) retains
  non-human structures for complete accounting. Category-level instance counts
  are deduplicated within each category; a physical instance can contribute to
  more than one category. The overall count and modality shares use the 13,826
  unique physical instances. Illustrations are cropped and
  arranged from \href{https://smart.servier.com/}{Servier Medical Art}, licensed
  under \href{https://creativecommons.org/licenses/by/4.0/}{CC BY 4.0}; they
  are not to scale and do not imply exhaustive organ coverage.}
  \label{fig:evaluation-benchmark-coverage}
\end{figure}

The benchmark covers major body regions, from the brain and head-and-neck
region to the thorax, abdomen, pelvis, spine, and extremities.
Representative coverage is shown in
\cref{fig:evaluation-benchmark-coverage}.

\paragraph{Training--test separation.}
None of the 4,326 test cases is used for training. Some source datasets also
contribute different cases to model development. The results therefore
measure performance on
unseen cases from a heterogeneous multi-source benchmark; they do not
establish transfer to entirely unseen datasets. We do not describe this
evaluation as external, zero-shot, fully independent, or prospective
validation. Dataset, institution, and case identifiers are withheld from the
public report.

\subsection{Interaction Modes and Recommended Use}
\label{sec:bbox-rationale}

A point prompt is quick to place, straightforward to interpret as indicating
foreground or background, and supported by all six evaluated methods.
Point-only prompting therefore provides an important common basis for
comparison. The first point prompt localizes a target but does not necessarily
identify it. This limitation is intrinsic to nested and adjacent structures:
a point inside a liver tumour,
for example, is simultaneously inside the tumour and the liver, so the
intended mask cannot be inferred from that coordinate alone.

A BBox supplies approximate extent and generally narrows the set of plausible
interpretations before refinement. It does not encode class identity or
eliminate every ambiguity, but it provides a more informative spatial
initialization across heterogeneous anatomical structures and pathological targets. We therefore
recommend BBox initialization followed by corrective points when approximate
extent is convenient to provide; point-only interaction remains the simpler
option and enables comparison with point-only systems such as VISTA3D.

\subsection{Simulated User Interaction Protocol}
\label{sec:simulated-interaction-protocol}

The point-only protocol adds five point prompts cumulatively. The first point
is placed at a maximum of the Euclidean distance transform (EDT) of the target
mask; subsequent points are selected from the largest 3D error component,
following the correction rule in the
\href{https://www.codabench.org/competitions/5263/}{CVPR 2025 Foundation
Models for Interactive 3D Biomedical Image Segmentation Challenge} protocol.
At each corrective round, the evaluator compares the current prediction with
the reference mask, selects the largest 26-connected error component, and places
a point at a maximally interior location determined by the EDT.
False-negative and false-positive regions yield positive and negative points,
respectively. Results are reported after one, three, and five points.
All methods use the same initial point prompts.
Subsequent point locations can differ because they are generated from each
model's current error mask.

For the BBox-initialized protocol, the simulator places a tight 2D box around
the largest connected component on the axial slice with the greatest target
area; no in-plane margin is added. It then performs five rounds of corrective
point prompting, as described in the point-only protocol. The BBox, preceding
points, and previous prediction are retained as inputs for the next round.

Every prompt is derived from a reference mask. The evaluation therefore
models an error-aware annotator operating under controlled conditions, with
each corrective point placed at an EDT maximum within the largest error
component. This simplified protocol does not reproduce the imprecision,
stopping decisions, timing, or variability of real users.

\subsection{Models and Protocol Compatibility}
\label{sec:evaluation-models}

The complete paired comparison in each interaction mode is between
\sami and nnInteractive, for which all 13,826 objects and all 107 categories
have valid inference results (\dice scores) at every stage.
The point-only table additionally reports
MedSAM2, SAM-Med3D, SegVol, and the VISTA3D CVPR 2025 checkpoint.
VISTA3D is included here because its published interactive branch accepts
positive and negative 3D points, but it is absent from the BBox table because
that branch does not define a BBox input compatible with this protocol.

MedSAM2, SAM-Med3D, and SegVol are included as additional baselines
under the BBox-plus-correction protocol.
They have valid scores for 13,420, 13,823, and 13,540 objects, respectively, at
each displayed stage. All three methods still cover all 107 categories.
Failed or invalid results are excluded rather than replaced with zero.
Method-specific coverage accompanies both comparisons
(\cref{tab:point-only-results,tab:bbox-point-results}).

It is worth noting that the baseline papers report different inference
procedures and prompt configurations: MedSAM2 initializes segmentation with
a box on the target's middle slice and propagates bidirectionally, while SegVol's main comparison uses
box and text prompts together \citep{ma2025medsam2,du2024segvol}. Rankings
here are specific to the evaluated interaction modes rather than a summary
of every prompting option offered by each method. VISTA3D is included as an
additional point-only baseline. For overlapping targets such as the pancreas
and pancreatic tumours, its published formulation assumes that the target
class $x$ is known and uses this information in an ambiguity embedding
\citep{he2025vista3d}.

\subsection{Metrics and Aggregation}
\label{sec:evaluation-metrics}

The principal metric is \dice on the $[0,1]$ scale. For each method and stage,
we first average valid object scores within every category and then weight
the available category means equally. For point-only interaction, we report
category-macro \dice after one, three, and five points, making both immediate
response and fixed-budget refinement visible. For BBox-initialized
interaction, we report initialization and predictions after one, three, and
five corrections. We do not combine the two modes into an equal-effort
ranking.

We report category-macro \dice scores stratified by object size, voxel
spacing, and imaging modality (\cref{sec:subgroup-results}). Within each subgroup, the comparison
uses the categories with at least one valid score at each of the BBox and
$+5$ stages for all five methods, so every method in a table row has the same
category denominator.
Size is defined by reference-mask voxel count (small: $<10{,}000$; medium:
10,000--99,999; large: $\geq100{,}000$). Acquisition spacing is the maximum
voxel spacing across three image axes (thin: $<1$ mm; intermediate: $[1,5)$
mm; thick: $\geq5$ mm). Modality comes from the audited object metadata
and target mappings.

\FloatBarrier
\section{Main Results}
\label{sec:main-results}

We first compare segmentation accuracy under point-only and box-initialized
interaction, then examine differences across medical categories and imaging
strata. The paired comparison with nnInteractive covers all 13,826 objects
and 107 categories. Additional baselines are compared under the evaluated
configurations, with method-specific coverage of valid scores reported
alongside the results
(\cref{sec:evaluation-models,tab:point-only-results,tab:bbox-point-results}).

\subsection{Point-Only Segmentation}
\label{sec:point-only-results}

\sami achieves the highest observed category-macro \dice among the evaluated
point-only configurations (\cref{tab:point-only-results}). It leads
nnInteractive, clearly the strongest baseline, from the first point, scoring
0.5756 versus 0.5316, and reaches 0.7771 versus 0.7495 with five total points.

\begin{table}[!htbp]
  \centering
  \caption{Point-only prompt comparison. Scores are category-macro \dice;
  invalid object scores are excluded rather than replaced with zero.
  VISTA3D's published formulation incorporates target-class information
  in addition to spatial prompts.}
  \label{tab:point-only-results}
  \small
  \setlength{\tabcolsep}{4.5pt}
  \begin{tabular}{lcccrr}
    \toprule
    Method & 1 point & 3 points & 5 points & Valid objects & Categories \\
    \midrule
    \textbf{\sami} & \textbf{0.5756} & \textbf{0.7396} & \textbf{0.7771} & 13,826 & 107 \\
    nnInteractive & 0.5316 & 0.6917 & 0.7495 & 13,826 & 107 \\
    MedSAM2 & 0.2790 & 0.3236 & 0.3657 & 13,715 & 107 \\
    SAM-Med3D & 0.3409 & 0.3930 & 0.4057 & 13,823 & 107 \\
    SegVol & 0.2774 & 0.2972 & 0.3138 & 13,785 & 107 \\
    VISTA3D (CVPR'25) & 0.3692 & 0.4185 & 0.4297 & 13,747 & 107 \\
    \bottomrule
  \end{tabular}
\end{table}

\FloatBarrier
\subsection{Box-Initialized Segmentation and Refinement}
\label{sec:bbox-point-results}

With box initialization, \sami has the highest observed category-macro
\dice among the five methods evaluated under this protocol
(\cref{tab:bbox-point-results}). Its advantage over nnInteractive is largest
before correction and narrows as corrective points are added. At
initialization, it scores 0.7129 versus 0.6530; after five corrections, the
scores are 0.8004 versus 0.7868. These accuracy trajectories describe distinct
prompting interfaces; the box and point budgets are not matched for user
effort (\cref{sec:bbox-rationale}).

\begin{table}[!htbp]
  \centering
  \caption{BBox-initialized point-refinement comparison. Scores are
  category-macro \dice computed from each method's valid object
  scores. Object and category coverage is identical across the four displayed
  stages; invalid scores are excluded, not replaced with zero.}
  \label{tab:bbox-point-results}
  \small
  \setlength{\tabcolsep}{4.2pt}
  \begin{tabular}{lccccrr}
    \toprule
    Method & BBox & $+1$ & $+3$ & $+5$ & Valid objects & Categories \\
    \midrule
    \textbf{\sami} & \textbf{0.7129} & \textbf{0.7605} & \textbf{0.7854} & \textbf{0.8004} & 13,826 & 107 \\
    nnInteractive & 0.6530 & 0.7153 & 0.7639 & 0.7868 & 13,826 & 107 \\
    MedSAM2 & 0.3764 & 0.3779 & 0.4101 & 0.4352 & 13,420 & 107 \\
    SAM-Med3D & 0.3404 & 0.3792 & 0.4009 & 0.4077 & 13,823 & 107 \\
    SegVol & 0.2402 & 0.2522 & 0.2578 & 0.2612 & 13,540 & 107 \\
    \bottomrule
  \end{tabular}
\end{table}

\FloatBarrier
\subsection{Category-Level Results}
\label{sec:category-results}
\label{sec:domain-results}

Detailed category comparisons focus on nnInteractive, for which complete
paired results are available. \Cref{tab:anatomy-category-results,tab:pathology-category-results}
list all 63 categories of anatomical structures and 43 pathology categories, with their full names,
object counts, and both methods' \dice at BBox initialization and after five
corrective points. Detailed results for \sami at BBox, $+1$, $+3$, and $+5$
appear in \cref{app:medical-target-evaluation}, using Combined Targets that
pool variants of each listed medical object across modalities and sides,
while retaining distinctions such as rib number and vertebral level.

\paragraph{Vascular targets and the renal collecting system.}
The largest final gains among anatomical structures occur in pulmonary vessels,
head and lower-limb vascular trees, and the renal collecting
system (\cref{tab:anatomy-category-results}). For head vascular trees,
\sami reaches 0.8228 versus 0.4455 after five corrections; the renal
collecting system improves by $+0.2299$. Neck vascular trees also benefit,
reaching 0.9576 versus 0.8805 over 50 objects. Pulmonary vessels, lower-limb
vascular trees, and the renal collecting system are represented by only
10, 11, and 20 objects, respectively. The large gains in these categories
are based on small samples and need confirmation in larger cohorts.

\paragraph{Gains across neoplastic categories.}
Pelvic, gastrointestinal, and bone tumors show final BBox-refinement gains
of $+0.2693$, $+0.1492$, and $+0.1527$, respectively
(\cref{tab:pathology-category-results}).

NF1-related tumors can occur throughout the body and exhibit complex
morphologies, including nodular, diffuse, and mixed forms. In this category
(381 objects from 31 cases), \sami achieves 0.6441 \dice at BBox
initialization, compared with 0.5512 for nnInteractive ($+0.0929$).
After five corrective clicks, \sami reaches 0.7351 versus 0.7077,
retaining an advantage of $+0.0275$.

\paragraph{Observed Relative Weaknesses.}
Brain favors \sami at five total points ($+0.1187$) but nnInteractive after
BBox refinement ($-0.1282$). Jaw bones show the largest final deficit among anatomical structures
with a box ($-0.6104$, 14 objects), despite an approximate tie in point-only
mode. CNS demyelinating and immune-mediated lesions favor nnInteractive in
both modes. No multiplicity-adjusted inference is made for these category-level
comparisons, and the means alone do not explain
failure mechanisms or establish vascular branch completeness.

\clearpage
\begingroup
\fontsize{8}{8.2}\selectfont
\setlength{\tabcolsep}{3pt}
\renewcommand{\arraystretch}{1.0}
\setlength{\LTleft}{\fill}
\setlength{\LTright}{\fill}
\setlength{\LTpre}{0pt}
\setlength{\LTpost}{0pt}
\begin{longtable}{@{}>{\raggedright\arraybackslash}p{0.18\textwidth}>{\raggedright\arraybackslash}p{0.37\textwidth}rcccc@{}}
\caption{Anatomical and Physiological Structures: BBox and BBox + 5 Clicks (Dice)}\label{tab:anatomy-category-results} \\
\toprule
Taxonomy & Category & \shortstack{Samples\\(instances)} & \multicolumn{2}{c}{\sami} & \multicolumn{2}{c}{nnInteractive} \\
\cmidrule(lr){4-5}\cmidrule(lr){6-7}
 &  &  & BBox & $+5$ & BBox & $+5$ \\
\midrule
\endfirsthead
\multicolumn{7}{c}{Table \thetable\ (continued)} \\
\toprule
Taxonomy & Category & \shortstack{Samples\\(instances)} & \multicolumn{2}{c}{\sami} & \multicolumn{2}{c}{nnInteractive} \\
\cmidrule(lr){4-5}\cmidrule(lr){6-7}
 &  &  & BBox & $+5$ & BBox & $+5$ \\
\midrule
\endhead
\midrule
\multicolumn{7}{r}{Continued on next page} \\
\endfoot
\bottomrule
\endlastfoot
 & \textbf{Mean} &  & \textbf{0.6862} & \textbf{0.7914} & 0.6321 & 0.7784 \\
\midrule
\multirow[t]{4}{=}{\raggedright \textbf{A01}\newline Nervous System} & Brain & 72 & 0.6170 & 0.6767 & \textbf{0.7260} & \textbf{0.8049} \\*
 & Ventricular \& Cerebrospinal Fluid System & 22 & \textbf{0.2870} & \textbf{0.4918} & 0.1857 & 0.3331 \\*
 & Spinal Cord & 89 & \textbf{0.6410} & \textbf{0.8487} & 0.6227 & 0.8381 \\*
 & Cranial Nerves & 41 & \textbf{0.4998} & \textbf{0.6043} & 0.3197 & 0.4840 \\
\addlinespace[1pt]
\multirow[t]{15}{=}{\raggedright \textbf{A02}\newline Cardiovascular System} & Heart & 119 & \textbf{0.8257} & 0.8545 & 0.8011 & \textbf{0.8830} \\*
 & Coronary Arterial Tree & 50 & 0.8152 & \textbf{0.9293} & \textbf{0.8555} & 0.9113 \\*
 & Aortic Vascular Tree & 29 & \textbf{0.8252} & \textbf{0.9173} & 0.6289 & 0.8397 \\*
 & Neck Vascular Tree & 50 & \textbf{0.8657} & \textbf{0.9576} & 0.7659 & 0.8805 \\*
 & Head Vascular Tree & 36 & \textbf{0.6611} & \textbf{0.8228} & 0.0747 & 0.4455 \\*
 & Pulmonary Circulation Vessels & 29 & \textbf{0.7214} & 0.8265 & 0.4141 & \textbf{0.8346} \\*
 & Pulmonary Vessels & 10 & \textbf{0.5637} & \textbf{0.8288} & 0.1181 & 0.3597 \\*
 & Hepatic Arteries and Veins & 24 & \textbf{0.8088} & \textbf{0.8518} & 0.7026 & 0.7984 \\*
 & Portal Venous System & 48 & \textbf{0.6355} & 0.7598 & 0.5167 & \textbf{0.7697} \\*
 & Hepatic Vessels & 16 & \textbf{0.2849} & \textbf{0.5908} & 0.2455 & 0.5367 \\*
 & Lower-limb Vascular Tree & 11 & \textbf{0.4800} & \textbf{0.7672} & 0.1854 & 0.5057 \\*
 & Brachiocephalic Veins & 68 & \textbf{0.8186} & 0.8987 & 0.8073 & \textbf{0.9202} \\*
 & Vena Caval System & 156 & \textbf{0.7454} & 0.8746 & 0.6124 & \textbf{0.8911} \\*
 & Other Arterial Vessels & 400 & \textbf{0.6981} & 0.8601 & 0.6855 & \textbf{0.8806} \\*
 & Other Venous Vessels & 61 & \textbf{0.6526} & 0.8323 & 0.6284 & \textbf{0.8505} \\
\addlinespace[1pt]
\multirow[t]{2}{=}{\raggedright \textbf{A03}\newline Respiratory System} & Respiratory Tract and Airway & 76 & \textbf{0.6650} & \textbf{0.7730} & 0.5933 & 0.7629 \\*
 & Lung & 387 & \textbf{0.9061} & 0.9290 & 0.8887 & \textbf{0.9465} \\
\addlinespace[1pt]
\multirow[t]{9}{=}{\raggedright \textbf{A04}\newline Digestive System} & Oral Cavity \& Pharynx & 42 & \textbf{0.6106} & \textbf{0.7039} & 0.5165 & 0.6936 \\*
 & Salivary Glands & 56 & \textbf{0.7905} & 0.8497 & 0.7521 & \textbf{0.8528} \\*
 & Esophagus & 146 & \textbf{0.5984} & 0.8076 & 0.5918 & \textbf{0.8213} \\*
 & Stomach & 130 & 0.8050 & 0.8800 & \textbf{0.8050} & \textbf{0.9035} \\*
 & Small Intestine & 224 & \textbf{0.5992} & 0.7458 & 0.5480 & \textbf{0.7678} \\*
 & Large Intestine & 207 & \textbf{0.5588} & 0.7431 & 0.5362 & \textbf{0.7729} \\*
 & Liver & 151 & 0.9284 & 0.9335 & \textbf{0.9319} & \textbf{0.9546} \\*
 & Gallbladder \& Biliary System & 121 & \textbf{0.7798} & 0.8453 & 0.7711 & \textbf{0.8602} \\*
 & Pancreas & 186 & \textbf{0.7382} & \textbf{0.8149} & 0.6492 & 0.8052 \\
\addlinespace[1pt]
\multirow[t]{4}{=}{\raggedright \textbf{A05}\newline Urinary System} & Kidney & 349 & \textbf{0.9279} & \textbf{0.9404} & 0.8829 & 0.9151 \\*
 & Renal Collecting System & 20 & \textbf{0.7158} & \textbf{0.7766} & 0.3857 & 0.5467 \\*
 & Ureter & 18 & \textbf{0.6432} & \textbf{0.7711} & 0.3397 & 0.6369 \\*
 & Urinary Bladder & 94 & \textbf{0.8111} & 0.8361 & 0.8062 & \textbf{0.8482} \\
\addlinespace[1pt]
\multirow[t]{2}{=}{\raggedright \textbf{A06}\newline Reproductive System} & Prostate & 89 & \textbf{0.7006} & 0.7782 & 0.6743 & \textbf{0.7945} \\*
 & Uterus \& Cervix & 8 & \textbf{0.8095} & \textbf{0.8916} & 0.7941 & 0.8824 \\
\addlinespace[1pt]
\multirow[t]{3}{=}{\raggedright \textbf{A07}\newline Endocrine System} & Pituitary Gland & 14 & \textbf{0.7247} & \textbf{0.7816} & 0.7075 & 0.7732 \\*
 & Thyroid Gland & 45 & \textbf{0.7059} & 0.8337 & 0.6523 & \textbf{0.8448} \\*
 & Adrenal Gland & 237 & \textbf{0.7391} & 0.7924 & 0.7308 & \textbf{0.8160} \\
\addlinespace[1pt]
\textbf{A08}\newline Lymphatic, Immune \& Hematopoietic System & Lymph Nodes & 103 & \textbf{0.7035} & 0.7660 & 0.6838 & \textbf{0.7818} \\*
 & Spleen & 129 & \textbf{0.9271} & 0.9380 & 0.9239 & \textbf{0.9473} \\
\addlinespace[1pt]
\multirow[t]{14}{=}{\raggedright \textbf{A09}\newline Skeletal \& Articular System} & Skull & 20 & 0.5389 & 0.7919 & \textbf{0.7700} & \textbf{0.8359} \\*
 & Jaw Bones & 14 & 0.0404 & 0.2028 & \textbf{0.6107} & \textbf{0.8132} \\*
 & Cervical Vertebrae & 117 & 0.6829 & 0.8021 & \textbf{0.8660} & \textbf{0.9114} \\*
 & Thoracic Vertebrae & 372 & 0.9114 & 0.9359 & \textbf{0.9315} & \textbf{0.9518} \\*
 & Lumbar Vertebrae & 118 & 0.9049 & 0.9273 & \textbf{0.9347} & \textbf{0.9583} \\*
 & Spinal Column, Spinal Canal and Level-unspecified Vertebra & 145 & \textbf{0.6140} & \textbf{0.7692} & 0.5225 & 0.7352 \\*
 & Intervertebral Disc & 182 & \textbf{0.8028} & \textbf{0.8190} & 0.7593 & 0.8173 \\*
 & Ribs & 779 & 0.8498 & 0.9121 & \textbf{0.8924} & \textbf{0.9363} \\*
 & Sternum & 37 & \textbf{0.8408} & 0.9247 & 0.6633 & \textbf{0.9351} \\*
 & Shoulder-girdle Bones & 139 & 0.8627 & 0.9074 & \textbf{0.8861} & \textbf{0.9169} \\*
 & Bony Pelvis & 118 & \textbf{0.8038} & 0.8763 & 0.7771 & \textbf{0.8795} \\*
 & Limb Bones & 335 & \textbf{0.8677} & \textbf{0.8945} & 0.8036 & 0.8937 \\*
 & Cartilage \& Fibrocartilage & 336 & \textbf{0.4218} & 0.6184 & 0.4180 & \textbf{0.6611} \\*
 & Ligament & 101 & \textbf{0.3348} & \textbf{0.4950} & 0.2337 & 0.4022 \\
\addlinespace[1pt]
\multirow[t]{6}{=}{\raggedright \textbf{A10}\newline Muscular System} & Head and Neck Muscles & 13 & \textbf{0.6187} & \textbf{0.7473} & 0.4063 & 0.7293 \\*
 & Abdominal Muscles & 92 & \textbf{0.7339} & 0.8480 & 0.7227 & \textbf{0.8703} \\*
 & Paraspinal Muscles & 125 & \textbf{0.7365} & 0.8726 & 0.7193 & \textbf{0.8779} \\*
 & Lower-limb Muscles & 180 & 0.7622 & 0.8659 & \textbf{0.7892} & \textbf{0.8665} \\*
 & Tendons \& Aponeuroses & 39 & \textbf{0.4233} & 0.5355 & 0.4104 & \textbf{0.5632} \\*
 & Fascia \& Muscular Compartments & 13 & 0.1836 & 0.2153 & \textbf{0.2003} & \textbf{0.2207} \\
\addlinespace[1pt]
\multirow[t]{2}{=}{\raggedright \textbf{A12}\newline Sensory System} & Eye \& Ocular Adnexa & 84 & \textbf{0.7420} & \textbf{0.8014} & 0.7249 & 0.7813 \\*
 & Ear \& Auditory-Vestibular Structures & 28 & 0.7173 & 0.7732 & \textbf{0.7192} & \textbf{0.7899} \\
\end{longtable}
\noindent\parbox{\textwidth}{\scriptsize\raggedright CT/MR cohort; taxonomy codes follow Table~\ref{tab:testset-taxonomy}. BBox is the initial 2D box; $+5$ denotes five corrective points. Category scores average valid instance Dice, counting each physical object once per category. Mean weights categories equally. Missing scores are excluded. Bold marks the higher score at the same stage.}\par
\endgroup

\clearpage
\begingroup
\fontsize{8}{8.2}\selectfont
\setlength{\tabcolsep}{3pt}
\renewcommand{\arraystretch}{1.0}
\setlength{\LTleft}{\fill}
\setlength{\LTright}{\fill}
\setlength{\LTpre}{0pt}
\setlength{\LTpost}{0pt}
\begin{longtable}{@{}>{\raggedright\arraybackslash}p{0.18\textwidth}>{\raggedright\arraybackslash}p{0.37\textwidth}rcccc@{}}
\caption{Pathological Findings and Abnormalities: BBox and BBox + 5 Clicks (Dice)}\label{tab:pathology-category-results} \\
\toprule
Taxonomy & Category & \shortstack{Samples\\(instances)} & \multicolumn{2}{c}{\sami} & \multicolumn{2}{c}{nnInteractive} \\
\cmidrule(lr){4-5}\cmidrule(lr){6-7}
 &  &  & BBox & $+5$ & BBox & $+5$ \\
\midrule
\endfirsthead
\multicolumn{7}{c}{Table \thetable\ (continued)} \\
\toprule
Taxonomy & Category & \shortstack{Samples\\(instances)} & \multicolumn{2}{c}{\sami} & \multicolumn{2}{c}{nnInteractive} \\
\cmidrule(lr){4-5}\cmidrule(lr){6-7}
 &  &  & BBox & $+5$ & BBox & $+5$ \\
\midrule
\endhead
\midrule
\multicolumn{7}{r}{Continued on next page} \\
\endfoot
\bottomrule
\endlastfoot
 & \textbf{Mean} &  & \textbf{0.7545} & \textbf{0.8185} & 0.6775 & 0.7954 \\
\midrule
\multirow[t]{23}{=}{\raggedright \textbf{B01}\newline Neoplastic Lesion} & Brain and Intracranial Tumor & 318 & \textbf{0.7942} & \textbf{0.8340} & 0.6950 & 0.7801 \\*
 & Parotid Tumor & 33 & \textbf{0.8381} & \textbf{0.8686} & 0.7922 & 0.8662 \\*
 & Lung Cancer or Pulmonary Mass & 238 & \textbf{0.7213} & \textbf{0.7913} & 0.6599 & 0.7798 \\*
 & Thymic Cancer & 70 & \textbf{0.7540} & \textbf{0.8380} & 0.6504 & 0.7475 \\*
 & Breast Tumor & 390 & \textbf{0.8039} & \textbf{0.8493} & 0.7217 & 0.8313 \\*
 & Gastric Cancer & 206 & \textbf{0.7046} & \textbf{0.7763} & 0.5376 & 0.6940 \\*
 & Gastrointestinal Stromal Tumor & 31 & \textbf{0.8255} & \textbf{0.8743} & 0.7737 & 0.8535 \\*
 & Gastrointestinal Tumor & 151 & \textbf{0.6481} & \textbf{0.7473} & 0.3679 & 0.5981 \\*
 & Liver Tumor & 339 & \textbf{0.7050} & \textbf{0.8012} & 0.7049 & 0.7969 \\*
 & Pancreatic Tumor & 94 & \textbf{0.8430} & \textbf{0.8832} & 0.7665 & 0.8732 \\*
 & Adrenal Tumor & 51 & 0.8268 & \textbf{0.9320} & \textbf{0.9000} & 0.9294 \\*
 & Kidney Cancer & 39 & \textbf{0.8295} & \textbf{0.8747} & 0.7200 & 0.8454 \\*
 & Kidney Tumor & 126 & \textbf{0.8931} & \textbf{0.9228} & 0.8566 & 0.9162 \\*
 & Prostate Tumor & 14 & \textbf{0.7311} & 0.7847 & 0.6673 & \textbf{0.8057} \\*
 & Cervical Cancer & 67 & \textbf{0.7505} & \textbf{0.7977} & 0.6441 & 0.7694 \\*
 & Pelvic Tumor & 64 & \textbf{0.7687} & \textbf{0.8251} & 0.4528 & 0.5558 \\*
 & Bone Tumor & 41 & \textbf{0.7372} & \textbf{0.8377} & 0.5183 & 0.6849 \\*
 & Soft-tissue Tumor & 48 & \textbf{0.7896} & 0.8383 & 0.7741 & \textbf{0.8469} \\*
 & Neurofibromatosis Type 1 (NF1) & 381 & \textbf{0.6441} & \textbf{0.7351} & 0.5512 & 0.7077 \\*
 & Metastatic Tumor & 404 & \textbf{0.7366} & \textbf{0.8178} & 0.7354 & 0.8096 \\*
 & Other Head and Neck Tumor & 167 & \textbf{0.7219} & 0.7986 & 0.7147 & \textbf{0.8362} \\*
 & Other Abdominal Tumor & 10 & \textbf{0.7861} & \textbf{0.8488} & 0.7089 & 0.8247 \\*
 & Other or Site-unspecified Tumor & 297 & \textbf{0.7316} & 0.8032 & 0.6848 & \textbf{0.8231} \\
\addlinespace[1pt]
\textbf{B02}\newline Infectious Lesion & Pneumonia & 428 & \textbf{0.6880} & 0.7778 & 0.6567 & \textbf{0.7836} \\
\addlinespace[1pt]
\textbf{B03}\newline Non-infectious Inflammatory \& Immune-mediated Lesion & Central Nervous-system Demyelinating and Immune-mediated Lesion & 149 & 0.7420 & 0.7363 & \textbf{0.7531} & \textbf{0.8566} \\
\addlinespace[1pt]
\textbf{B04}\newline Vascular \& Circulatory Disorder & Intracranial Hemorrhage & 461 & \textbf{0.6421} & \textbf{0.7051} & 0.5823 & 0.6870 \\*
 & Cerebral Ischemia, Infarction and Hypoxic-ischemic Injury & 173 & \textbf{0.6256} & 0.7130 & 0.6082 & \textbf{0.7407} \\
\addlinespace[1pt]
\textbf{B05}\newline Trauma \& Tissue Injury & Fracture & 184 & \textbf{0.9279} & \textbf{0.9606} & 0.8935 & 0.9491 \\*
 & Anterior Cruciate Ligament Injury & 54 & \textbf{0.7102} & \textbf{0.7557} & 0.5101 & 0.6724 \\
\addlinespace[1pt]
\textbf{B07}\newline Non-neoplastic Focal Lesion & Renal Cyst & 84 & \textbf{0.8961} & \textbf{0.9258} & 0.8751 & 0.9230 \\
\addlinespace[1pt]
\multirow[t]{13}{=}{\raggedright \textbf{B10}\newline Indeterminate Imaging Abnormality} & Indeterminate Nervous-system Imaging Abnormality & 20 & \textbf{0.7854} & \textbf{0.8550} & 0.7005 & 0.7963 \\*
 & Thyroid Nodule & 30 & \textbf{0.7753} & \textbf{0.8326} & 0.6685 & 0.8043 \\*
 & Pulmonary Nodule & 405 & \textbf{0.8029} & \textbf{0.8493} & 0.7616 & 0.8387 \\*
 & Breast Nodule & 76 & \textbf{0.8424} & \textbf{0.8696} & 0.7516 & 0.8507 \\*
 & Indeterminate Breast Lesion & 77 & \textbf{0.7067} & \textbf{0.7823} & 0.5649 & 0.7427 \\*
 & Indeterminate Liver Lesion & 121 & \textbf{0.7687} & \textbf{0.8432} & 0.7488 & 0.8343 \\*
 & Indeterminate Pancreatic Lesion & 12 & \textbf{0.7402} & 0.7850 & 0.6387 & \textbf{0.8164} \\*
 & Indeterminate Prostate Lesion & 91 & 0.6363 & 0.7476 & \textbf{0.6590} & \textbf{0.7724} \\*
 & Indeterminate Ovarian Lesion & 87 & \textbf{0.8264} & 0.8669 & 0.7922 & \textbf{0.8864} \\*
 & Perirectal Lymph-node Target of Indeterminate Status & 15 & \textbf{0.8573} & \textbf{0.8806} & 0.5408 & 0.7826 \\*
 & Indeterminate Lymph-node Target & 188 & \textbf{0.6914} & 0.7643 & 0.6817 & \textbf{0.7838} \\*
 & Pathologic Fluid Collection and Abnormal Gas in Body Cavity & 15 & \textbf{0.4129} & 0.6253 & 0.4126 & \textbf{0.6734} \\*
 & Other Indeterminate Abdominopelvic Lesion & 145 & \textbf{0.7810} & \textbf{0.8376} & 0.7351 & 0.8303 \\
\end{longtable}
\noindent\parbox{\textwidth}{\scriptsize\raggedright CT/MR cohort; taxonomy codes follow Table~\ref{tab:testset-taxonomy}. BBox is the initial 2D box; $+5$ denotes five corrective points. Category scores average valid instance Dice, counting each physical object once per category. Mean weights categories equally. Missing scores are excluded. Bold marks the higher score at the same stage.}\par
\endgroup

\clearpage

\subsection{Stratified Results}
\label{sec:subgroup-results}
\label{sec:modality-results}

\paragraph{Imaging modality.}
\sami has the highest observed BBox-initialized category-macro score in
both evaluated modalities. After five corrections, it remains highest on CT
(0.8302 versus 0.8195 for nnInteractive) and MR (0.7455 versus 0.7433)
(\cref{tab:modality-subgroups}). CT and MR contribute 85 and 69 represented
categories, respectively; these sets overlap and differ in target composition.

\begin{table}[!htbp]
  \centering
  \caption{Category-macro \dice by 3D imaging modality. Each method cell
  reports BBox / $+5$ corrections. Categories denotes the common denominator
  of categories with valid scores for all five methods; bold marks the best
  value at each stage within a row. The cohort is restricted to CT and MR.}
  \label{tab:modality-subgroups}
  \scriptsize
  \setlength{\tabcolsep}{3.2pt}
  \renewcommand{\arraystretch}{1.18}
  \resizebox{\textwidth}{!}{%
  \begin{tabular}{lrrccccc}
    \toprule
    Modality & Categories (used/mapped) & Objects & \sami & nnInteractive & MedSAM2 & SAM-Med3D & SegVol \\
    \midrule
    CT & 85/85 & 9,289 & \textbf{0.7445} / \textbf{0.8302} & 0.6870 / 0.8195 & 0.3769 / 0.4161 & 0.3578 / 0.4367 & 0.2285 / 0.2497 \\
    MR & 69/69 & 4,537 & \textbf{0.6405} / \textbf{0.7455} & 0.5997 / 0.7433 & 0.3384 / 0.4080 & 0.3378 / 0.4162 & 0.2382 / 0.2741 \\
    \bottomrule
  \end{tabular}%
  }
\end{table}

\paragraph{Object size and voxel spacing.}
\sami has the highest initial and final category-macro \dice in each
reported size and maximum-spacing stratum. The final margins over
nnInteractive are small: $+0.0034$ to $+0.0099$ across size strata and
$+0.0059$ to $+0.0178$ across spacing strata. Full five-method results and
category denominators appear in
\cref{tab:instance-size-subgroups,tab:max-spacing-subgroups}.
These exploratory comparisons use common category denominators based on
valid scores within each stratum, but do not control for differences in target composition or
support causal claims about size, spacing, or modality.

\begin{table}[!htbp]
  \centering
  \caption{Category-macro \dice by reference-object size. Each method cell
  reports BBox / $+5$ corrections. Categories denotes the common denominator
  of categories with valid scores for all five methods; bold marks the best
  value at each stage within a row.}
  \label{tab:instance-size-subgroups}
  \scriptsize
  \setlength{\tabcolsep}{3.2pt}
  \renewcommand{\arraystretch}{1.18}
  \resizebox{\textwidth}{!}{%
  \begin{tabular}{lrrccccc}
    \toprule
    Size by target voxels & Categories (used/mapped) & Objects & \sami & nnInteractive & MedSAM2 & SAM-Med3D & SegVol \\
    \midrule
    Small ($<10{,}000$) & 97/97 & 7,855 & \textbf{0.6746} / \textbf{0.7650} & 0.6123 / 0.7551 & 0.3638 / 0.4728 & 0.2939 / 0.3223 & 0.3428 / 0.3589 \\
    Medium ($10{,}000$--$99{,}999$) & 90/90 & 4,009 & \textbf{0.7443} / \textbf{0.8270} & 0.7089 / 0.8231 & 0.4134 / 0.4285 & 0.4400 / 0.5255 & 0.1644 / 0.2127 \\
    Large ($\geq100{,}000$) & 68/69 & 1,962 & \textbf{0.7523} / \textbf{0.8392} & 0.7237 / 0.8358 & 0.4782 / 0.4181 & 0.4711 / 0.6068 & 0.0555 / 0.1128 \\
    \bottomrule
  \end{tabular}%
  }
\end{table}

\begin{table}[!htbp]
  \centering
  \caption{Category-macro \dice by maximum voxel spacing. Each method cell
  reports BBox / $+5$ corrections. Categories denotes the common denominator
  of categories with valid scores for all five methods; bold marks the best
  value at each stage within a row.}
  \label{tab:max-spacing-subgroups}
  \scriptsize
  \setlength{\tabcolsep}{3.2pt}
  \renewcommand{\arraystretch}{1.18}
  \resizebox{\textwidth}{!}{%
  \begin{tabular}{lrrccccc}
    \toprule
    Maximum spacing & Categories (used/mapped) & Objects & \sami & nnInteractive & MedSAM2 & SAM-Med3D & SegVol \\
    \midrule
    Thin ($<1$ mm) & 46/47 & 981 & \textbf{0.7826} / \textbf{0.8403} & 0.7282 / 0.8343 & 0.3616 / 0.3402 & 0.3810 / 0.4828 & 0.1127 / 0.1064 \\
    Intermediate ($1$--$<5$ mm) & 104/104 & 9,657 & \textbf{0.7168} / \textbf{0.8028} & 0.6518 / 0.7850 & 0.3637 / 0.4151 & 0.3370 / 0.4075 & 0.2204 / 0.2371 \\
    Thick ($\geq5$ mm) & 70/70 & 3,188 & \textbf{0.6803} / \textbf{0.7768} & 0.6338 / 0.7688 & 0.4425 / 0.5400 & 0.3851 / 0.4355 & 0.3421 / 0.3709 \\
    \bottomrule
  \end{tabular}%
  }
\end{table}

\FloatBarrier

\section{Limitations and Responsible Use}
\label{sec:limitations-responsible-use}

\paragraph{Evaluation scope.}
Training and test cases are disjoint, but some source datasets contribute to
both splits; transfer to unseen datasets and institutions remains untested.
The current evaluation covers CT and MR only. Future work will include PET,
3D ultrasound, and independent external validation.

\paragraph{Simulated interaction.}
Reference masks determine the targets and prompts, so this benchmark does not
measure human prompt variability or annotation time. The preliminary NF1
timing comparison is task-specific and requires fuller reporting. Predicted
masks and subsequent corrections require qualified human review.

\paragraph{Baseline comparability.}
Coverage of valid scores differs across baselines. VISTA3D's published
formulation also uses target-class information. These comparisons are
specific to the evaluated configurations. Category averages do not establish uniform
superiority across targets or clinical effectiveness.

\section{Conclusion}
\label{sec:conclusion}

We present \sami and evaluate interactive 3D segmentation on a CT/MR
benchmark of 219 source datasets, 4,326 cases, and 107 medical categories.
With equal category weighting, \sami achieves the highest observed mean
\dice among the evaluated configurations: 0.5756 with one point, 0.7771
with five points, 0.7129 with BBox initialization, and 0.8004 after five
corrective clicks. These results support point-only interaction as a simple
interface and BBox initialization when approximate target extent is
available. Independent external validation, real-user studies, and evaluation
on additional imaging modalities remain necessary to establish broader
applicability and workflow benefits.

\section*{Acknowledgments}
\phantomsection
\addcontentsline{toc}{section}{Acknowledgments}

We would like to thank Professor Zhichao Wang and Dr.\ Jun Liu, from Shanghai Jiao Tong
University School of Medicine and its affiliated Shanghai Ninth People's
Hospital, for their valuable feedback during the early development of \sami.

\clearpage
\bibliographystyle{abbrvnat}
\bibliography{references}

\clearpage
\appendix
\section{Detailed Medical Target Evaluation}
\label{app:medical-target-evaluation}

To present target-level results concisely, we pool \emph{Specific Targets}
for the same medical object across modalities and sides into
\emph{Combined Targets}. For example, left and right kidney targets from
CT and MR are combined into a single \emph{Kidney} row.

\begingroup
\fontsize{8}{9}\selectfont
\setlength{\tabcolsep}{3pt}
\renewcommand{\arraystretch}{1.12}
\setlength{\LTleft}{\fill}
\setlength{\LTright}{\fill}
\begin{longtable}{@{}>{\raggedright\arraybackslash}p{0.14\textwidth}>{\raggedright\arraybackslash}p{0.20\textwidth}>{\raggedright\arraybackslash}p{0.22\textwidth}rcccc@{}}
\caption{Anatomical and Physiological Targets: BBox + Point Segmentation (Dice)}\label{tab:anatomy_bbox_point_target_dice} \\
\toprule
Taxonomy & Category & Combined Targets & \shortstack{Samples\\(instances)} & \multicolumn{4}{c}{\sami} \\
\cmidrule(lr){5-8}
 & & & & BBox & $+1$ & $+3$ & $+5$ \\
\midrule
\endfirsthead
\multicolumn{8}{c}{Table \thetable\ (continued)} \\
\toprule
Taxonomy & Category & Combined Targets & \shortstack{Samples\\(instances)} & \multicolumn{4}{c}{\sami} \\
\cmidrule(lr){5-8}
 & & & & BBox & $+1$ & $+3$ & $+5$ \\
\midrule
\endhead
\midrule
\multicolumn{8}{r}{Continued on next page} \\
\endfoot
\bottomrule
\endlastfoot
\textbf{Mean} &  &  &  & 0.7044 & 0.7521 & 0.7770 & 0.7978 \\
\midrule
\textbf{A01 Nervous System} & Brain & Brain & 14 & 0.9121 & 0.9063 & 0.9174 & 0.9279 \\*
 &  & Brainstem & 14 & 0.7552 & 0.8059 & 0.8275 & 0.8416 \\*
 &  & Cerebral Gray Matter & 22 & 0.4740 & 0.6228 & 0.4513 & 0.4913 \\*
 &  & Cerebral White Matter & 22 & 0.4841 & 0.3970 & 0.5267 & 0.5973 \\
\addlinespace[2pt]
 & Ventricular \& Cerebrospinal Fluid System & Cerebrospinal Fluid & 22 & 0.2870 & 0.3118 & 0.4743 & 0.4918 \\
\addlinespace[2pt]
 & Spinal Cord & Spinal Cord & 89 & 0.6410 & 0.7856 & 0.8316 & 0.8487 \\
\addlinespace[2pt]
 & Cranial Nerves & Optic Nerve & 28 & 0.5084 & 0.5770 & 0.6204 & 0.6418 \\*
 &  & Optic Chiasm & 13 & 0.4811 & 0.4886 & 0.5142 & 0.5236 \\
\addlinespace[4pt]
\textbf{A02 Cardiovascular System} & Heart & Heart & 51 & 0.8642 & 0.8830 & 0.8888 & 0.8972 \\*
 &  & Left Atrium & 10 & 0.9302 & 0.9364 & 0.9400 & 0.9435 \\*
 &  & Left Ventricular Cavity & 10 & 0.8859 & 0.9104 & 0.9158 & 0.9195 \\*
 &  & Right Ventricular Cavity & 10 & 0.8073 & 0.8399 & 0.8586 & 0.8623 \\*
 &  & Myocardium & 10 & 0.4342 & 0.4637 & 0.4329 & 0.5267 \\*
 &  & Left Atrial Appendage & 28 & 0.8429 & 0.8288 & 0.8234 & 0.8360 \\
\addlinespace[2pt]
 & Coronary Arterial Tree & Coronary Arterial Tree & 50 & 0.8152 & 0.8500 & 0.9050 & 0.9293 \\
\addlinespace[2pt]
 & Aortic Vascular Tree & Aortic Vessel Tree & 29 & 0.8252 & 0.8968 & 0.9189 & 0.9173 \\
\addlinespace[2pt]
 & Neck Vascular Tree & Neck Arteries & 50 & 0.8657 & 0.9535 & 0.9600 & 0.9576 \\
\addlinespace[2pt]
 & Head Vascular Tree & Head Arteries & 36 & 0.6611 & 0.7902 & 0.8201 & 0.8228 \\
\addlinespace[2pt]
 & Pulmonary Circulation Vessels & Pulmonary Vein & 29 & 0.7214 & 0.8061 & 0.8178 & 0.8265 \\
\addlinespace[2pt]
 & Pulmonary Vessels & Pulmonary Vessels & 10 & 0.5637 & 0.7347 & 0.8170 & 0.8288 \\
\addlinespace[2pt]
 & Hepatic Arteries and Veins & Hepatic Artery & 12 & 0.8549 & 0.8739 & 0.8788 & 0.8857 \\*
 &  & Hepatic Vein & 12 & 0.7627 & 0.7788 & 0.7946 & 0.8180 \\
\addlinespace[2pt]
 & Portal Venous System & Portal Vein & 12 & 0.7301 & 0.7643 & 0.7342 & 0.7754 \\*
 &  & Portal and Splenic Vein Union & 36 & 0.6040 & 0.7125 & 0.7358 & 0.7547 \\
\addlinespace[2pt]
 & Hepatic Vessels & Hepatic Vessels & 16 & 0.2849 & 0.4818 & 0.5621 & 0.5908 \\
\addlinespace[2pt]
 & Lower-limb Vascular Tree & Lower-limb Arteries & 11 & 0.4800 & 0.5886 & 0.6836 & 0.7672 \\
\addlinespace[2pt]
 & Brachiocephalic Veins & Brachiocephalic Vein & 68 & 0.8186 & 0.8641 & 0.8880 & 0.8987 \\
\addlinespace[2pt]
 & Vena Caval System & Superior Vena Cava & 32 & 0.8622 & 0.9031 & 0.9167 & 0.9196 \\*
 &  & Inferior Vena Cava & 124 & 0.7153 & 0.8264 & 0.8471 & 0.8630 \\
\addlinespace[2pt]
 & Other Arterial Vessels & Aorta & 140 & 0.8141 & 0.8668 & 0.8887 & 0.9005 \\*
 &  & Common Carotid Artery & 94 & 0.5746 & 0.7321 & 0.8041 & 0.8346 \\*
 &  & Subclavian Artery & 69 & 0.6631 & 0.8024 & 0.8515 & 0.8678 \\*
 &  & Brachiocephalic Trunk & 34 & 0.8072 & 0.8332 & 0.8714 & 0.8836 \\*
 &  & Iliac Artery & 63 & 0.6043 & 0.6830 & 0.7329 & 0.7874 \\
\addlinespace[2pt]
 & Other Venous Vessels & Iliac Vein & 61 & 0.6526 & 0.7595 & 0.8098 & 0.8323 \\
\addlinespace[4pt]
\textbf{A03 Respiratory System} & Respiratory Tract and Airway & Trachea & 36 & 0.8526 & 0.9104 & 0.9206 & 0.9291 \\*
 &  & Glottis & 13 & 0.4361 & 0.4476 & 0.4783 & 0.5471 \\*
 &  & Supraglottic Larynx & 14 & 0.6307 & 0.6588 & 0.7115 & 0.7136 \\*
 &  & Arytenoid Cartilage & 13 & 0.4113 & 0.5602 & 0.6070 & 0.6304 \\
\addlinespace[2pt]
 & Lung & Lung & 204 & 0.9079 & 0.9312 & 0.9319 & 0.9344 \\*
 &  & Lung Lobe & 183 & 0.9041 & 0.9140 & 0.9229 & 0.9231 \\
\addlinespace[4pt]
\textbf{A04 Digestive System} & Oral Cavity \& Pharynx & Oral Cavity & 14 & 0.8531 & 0.8729 & 0.8824 & 0.8836 \\*
 &  & Buccal Mucosa & 14 & 0.4474 & 0.5591 & 0.5914 & 0.5915 \\*
 &  & Lips & 14 & 0.5314 & 0.5899 & 0.6288 & 0.6365 \\
\addlinespace[2pt]
 & Salivary Glands & Parotid Gland & 28 & 0.7945 & 0.8080 & 0.8193 & 0.8339 \\*
 &  & Submandibular Gland & 28 & 0.7866 & 0.8364 & 0.8557 & 0.8655 \\
\addlinespace[2pt]
 & Esophagus & Esophagus & 146 & 0.5984 & 0.7423 & 0.7922 & 0.8076 \\
\addlinespace[2pt]
 & Stomach & Stomach & 130 & 0.8050 & 0.8608 & 0.8803 & 0.8800 \\
\addlinespace[2pt]
 & Small Intestine & Duodenum & 122 & 0.6030 & 0.7138 & 0.7626 & 0.7816 \\*
 &  & Small Intestine & 102 & 0.5947 & 0.6611 & 0.6955 & 0.7029 \\
\addlinespace[2pt]
 & Large Intestine & Appendix & 5 & 0.3531 & 0.4432 & 0.4606 & 0.4813 \\*
 &  & Cecum & 10 & 0.6300 & 0.6736 & 0.7097 & 0.7218 \\*
 &  & Colon & 97 & 0.5538 & 0.6681 & 0.7467 & 0.7484 \\*
 &  & Ascending Colon & 10 & 0.7523 & 0.8036 & 0.7881 & 0.7940 \\*
 &  & Transverse Colon & 10 & 0.3130 & 0.3730 & 0.4901 & 0.5753 \\*
 &  & Descending Colon & 10 & 0.5422 & 0.6765 & 0.7365 & 0.7614 \\*
 &  & Sigmoid Colon & 10 & 0.2989 & 0.4344 & 0.6082 & 0.6523 \\*
 &  & Rectum & 55 & 0.6331 & 0.7429 & 0.7782 & 0.7958 \\
\addlinespace[2pt]
 & Liver & Liver & 151 & 0.9284 & 0.9348 & 0.9416 & 0.9335 \\
\addlinespace[2pt]
 & Gallbladder \& Biliary System & Gallbladder & 113 & 0.7993 & 0.8296 & 0.8478 & 0.8589 \\*
 &  & Bile Duct & 8 & 0.5042 & 0.5942 & 0.6218 & 0.6536 \\
\addlinespace[2pt]
 & Pancreas & Pancreas & 186 & 0.7382 & 0.7819 & 0.8048 & 0.8149 \\
\addlinespace[4pt]
\textbf{A05 Urinary System} & Kidney & Kidney & 309 & 0.9341 & 0.9427 & 0.9484 & 0.9497 \\*
 &  & Renal Cortex & 20 & 0.9060 & 0.9036 & 0.8249 & 0.8766 \\*
 &  & Renal Medulla & 20 & 0.8542 & 0.8592 & 0.8607 & 0.8615 \\
\addlinespace[2pt]
 & Renal Collecting System & Renal Pelvis and Calyces & 20 & 0.7158 & 0.7426 & 0.7620 & 0.7766 \\
\addlinespace[2pt]
 & Ureter & Ureter & 18 & 0.6432 & 0.7004 & 0.7444 & 0.7711 \\
\addlinespace[2pt]
 & Urinary Bladder & Urinary Bladder & 94 & 0.8111 & 0.8149 & 0.8308 & 0.8361 \\
\addlinespace[4pt]
\textbf{A06 Reproductive System} & Prostate & Prostate & 70 & 0.7200 & 0.7560 & 0.7798 & 0.7975 \\*
 &  & Prostate Peripheral Zone & 10 & 0.4955 & 0.5394 & 0.5777 & 0.5984 \\*
 &  & Prostate Transition Zone & 9 & 0.7774 & 0.7954 & 0.8029 & 0.8284 \\
\addlinespace[2pt]
 & Uterus \& Cervix & Uterus & 8 & 0.8095 & 0.8702 & 0.8735 & 0.8916 \\
\addlinespace[4pt]
\textbf{A07 Endocrine System} & Pituitary Gland & Pituitary Gland & 14 & 0.7247 & 0.7701 & 0.7813 & 0.7816 \\
\addlinespace[2pt]
 & Thyroid Gland & Thyroid Gland & 45 & 0.7059 & 0.7992 & 0.8291 & 0.8337 \\
\addlinespace[2pt]
 & Adrenal Gland & Adrenal Gland & 237 & 0.7391 & 0.7711 & 0.7877 & 0.7924 \\
\addlinespace[4pt]
\textbf{A08 Lymphatic, Immune \& Hematopoietic System} & Lymph Nodes & Lymph Node & 103 & 0.7035 & 0.7369 & 0.7638 & 0.7660 \\
\addlinespace[2pt]
 & Spleen & Spleen & 129 & 0.9271 & 0.9256 & 0.9326 & 0.9380 \\
\addlinespace[4pt]
\textbf{A09 Skeletal \& Articular System} & Skull & Skull & 20 & 0.5389 & 0.5824 & 0.6369 & 0.7919 \\
\addlinespace[2pt]
 & Jaw Bones & Mandible & 14 & 0.0404 & 0.1079 & 0.1252 & 0.2028 \\
\addlinespace[2pt]
 & Cervical Vertebrae & C1 Vertebra & 11 & 0.7951 & 0.8458 & 0.8647 & 0.8838 \\*
 &  & C2 Vertebra & 12 & 0.7914 & 0.7541 & 0.7659 & 0.8219 \\*
 &  & C3 Vertebra & 10 & 0.5887 & 0.4703 & 0.4697 & 0.7186 \\*
 &  & C4 Vertebra & 12 & 0.5029 & 0.4822 & 0.5243 & 0.7901 \\*
 &  & C5 Vertebra & 15 & 0.5221 & 0.5466 & 0.6156 & 0.6658 \\*
 &  & C6 Vertebra & 25 & 0.6277 & 0.6740 & 0.7239 & 0.7615 \\*
 &  & C7 Vertebra & 32 & 0.8189 & 0.8368 & 0.8919 & 0.8929 \\
\addlinespace[2pt]
 & Thoracic Vertebrae & T1 Vertebra & 32 & 0.9318 & 0.9401 & 0.9438 & 0.9464 \\*
 &  & T2 Vertebra & 33 & 0.8969 & 0.8807 & 0.9117 & 0.9282 \\*
 &  & T3 Vertebra & 34 & 0.8644 & 0.8615 & 0.8779 & 0.8991 \\*
 &  & T4 Vertebra & 33 & 0.8736 & 0.9118 & 0.9016 & 0.9161 \\*
 &  & T5 Vertebra & 32 & 0.8462 & 0.8462 & 0.8714 & 0.8841 \\*
 &  & T6 Vertebra & 29 & 0.9073 & 0.9144 & 0.9132 & 0.9209 \\*
 &  & T7 Vertebra & 27 & 0.9402 & 0.9341 & 0.9507 & 0.9551 \\*
 &  & T8 Vertebra & 28 & 0.9334 & 0.9533 & 0.9577 & 0.9580 \\*
 &  & T9 Vertebra & 30 & 0.9414 & 0.9593 & 0.9626 & 0.9635 \\*
 &  & T10 Vertebra & 31 & 0.9401 & 0.9562 & 0.9582 & 0.9612 \\*
 &  & T11 Vertebra & 32 & 0.9350 & 0.9498 & 0.9521 & 0.9524 \\*
 &  & T12 Vertebra & 31 & 0.9419 & 0.9477 & 0.9561 & 0.9568 \\
\addlinespace[2pt]
 & Lumbar Vertebrae & L1 Vertebra & 30 & 0.9582 & 0.9216 & 0.9483 & 0.9457 \\*
 &  & L2 Vertebra & 26 & 0.8810 & 0.9235 & 0.9285 & 0.9264 \\*
 &  & L3 Vertebra & 21 & 0.9059 & 0.9642 & 0.9652 & 0.9676 \\*
 &  & L4 Vertebra & 20 & 0.9368 & 0.8921 & 0.8766 & 0.9152 \\*
 &  & L5 Vertebra & 21 & 0.8268 & 0.7900 & 0.8565 & 0.8730 \\
\addlinespace[2pt]
 & Spinal Column, Spinal Canal and Level-unspecified Vertebra & Vertebra, Level Unspecified & 122 & 0.5892 & 0.7020 & 0.7386 & 0.7496 \\*
 &  & Vertebral Column & 23 & 0.7456 & 0.8518 & 0.8666 & 0.8730 \\
\addlinespace[2pt]
 & Intervertebral Disc & Intervertebral Disc & 182 & 0.8028 & 0.8109 & 0.8160 & 0.8190 \\
\addlinespace[2pt]
 & Ribs & Rib 1 & 70 & 0.8141 & 0.8553 & 0.8807 & 0.8910 \\*
 &  & Rib 2 & 70 & 0.9016 & 0.9193 & 0.9263 & 0.9251 \\*
 &  & Rib 3 & 71 & 0.8482 & 0.8713 & 0.8680 & 0.8747 \\*
 &  & Rib 4 & 66 & 0.8663 & 0.9090 & 0.9132 & 0.9222 \\*
 &  & Rib 5 & 68 & 0.8348 & 0.9015 & 0.8991 & 0.9093 \\*
 &  & Rib 6 & 63 & 0.8124 & 0.9043 & 0.8921 & 0.9037 \\*
 &  & Rib 7 & 60 & 0.8030 & 0.8740 & 0.8748 & 0.8966 \\*
 &  & Rib 8 & 64 & 0.8397 & 0.8973 & 0.9106 & 0.9219 \\*
 &  & Rib 9 & 62 & 0.8779 & 0.9199 & 0.9364 & 0.9342 \\*
 &  & Rib 10 & 61 & 0.8594 & 0.9176 & 0.9365 & 0.9357 \\*
 &  & Rib 11 & 64 & 0.8764 & 0.9146 & 0.9186 & 0.9250 \\*
 &  & Rib 12 & 60 & 0.8612 & 0.9012 & 0.9074 & 0.9113 \\
\addlinespace[2pt]
 & Sternum & Sternum & 37 & 0.8408 & 0.8954 & 0.9189 & 0.9247 \\
\addlinespace[2pt]
 & Shoulder-girdle Bones & Scapula & 68 & 0.8145 & 0.8632 & 0.8729 & 0.8782 \\*
 &  & Clavicle & 71 & 0.9088 & 0.9248 & 0.9300 & 0.9353 \\
\addlinespace[2pt]
 & Bony Pelvis & Sacrum & 33 & 0.8452 & 0.8515 & 0.8711 & 0.8736 \\*
 &  & Hip Bone & 65 & 0.7626 & 0.8065 & 0.8460 & 0.8712 \\*
 &  & S1 Vertebra & 20 & 0.8694 & 0.8847 & 0.8934 & 0.8972 \\
\addlinespace[2pt]
 & Limb Bones & Humerus & 59 & 0.8848 & 0.9012 & 0.9146 & 0.9171 \\*
 &  & Femur & 157 & 0.8866 & 0.8988 & 0.9044 & 0.9074 \\*
 &  & Femoral Head & 21 & 0.8736 & 0.8688 & 0.9051 & 0.9159 \\*
 &  & Tibia & 34 & 0.8763 & 0.8708 & 0.8676 & 0.8633 \\*
 &  & Fibula & 34 & 0.7726 & 0.8137 & 0.8210 & 0.8404 \\*
 &  & Patella & 30 & 0.8289 & 0.8452 & 0.8547 & 0.8646 \\
\addlinespace[2pt]
 & Cartilage \& Fibrocartilage & Costal Cartilage & 39 & 0.4945 & 0.5977 & 0.6483 & 0.6735 \\*
 &  & Meniscus & 102 & 0.4237 & 0.5862 & 0.6742 & 0.6862 \\*
 &  & Patellar Cartilage & 51 & 0.6030 & 0.7396 & 0.7875 & 0.7885 \\*
 &  & Femoral Articular Cartilage & 63 & 0.1429 & 0.2351 & 0.2685 & 0.3003 \\*
 &  & Tibial Articular Cartilage & 81 & 0.4871 & 0.5973 & 0.6218 & 0.6470 \\
\addlinespace[2pt]
 & Ligament & Anterior Cruciate Ligament & 20 & 0.5456 & 0.6292 & 0.6666 & 0.6881 \\*
 &  & Posterior Cruciate Ligament & 20 & 0.3739 & 0.4596 & 0.5032 & 0.5484 \\*
 &  & Patellar Ligament & 7 & 0.2775 & 0.2355 & 0.3582 & 0.3847 \\*
 &  & Knee Collateral Ligament & 26 & 0.2223 & 0.2462 & 0.2586 & 0.2756 \\*
 &  & Patellar Retinaculum & 28 & 0.2751 & 0.3169 & 0.4211 & 0.5502 \\
\addlinespace[4pt]
\textbf{A10 Muscular System} & Head and Neck Muscles & Cricopharyngeus Muscle & 13 & 0.6187 & 0.6939 & 0.7283 & 0.7473 \\
\addlinespace[2pt]
 & Abdominal Muscles & Iliopsoas & 92 & 0.7339 & 0.8188 & 0.8376 & 0.8480 \\
\addlinespace[2pt]
 & Paraspinal Muscles & Intrinsic Back Muscle & 125 & 0.7365 & 0.8286 & 0.8586 & 0.8726 \\
\addlinespace[2pt]
 & Lower-limb Muscles & Gluteus Maximus & 59 & 0.8458 & 0.8706 & 0.8817 & 0.8897 \\*
 &  & Gluteus Medius & 62 & 0.6933 & 0.8489 & 0.8625 & 0.8803 \\*
 &  & Gluteus Minimus & 59 & 0.7510 & 0.7859 & 0.8194 & 0.8269 \\
\addlinespace[2pt]
 & Tendons \& Aponeuroses & Quadriceps Tendon & 7 & 0.5419 & 0.5582 & 0.6181 & 0.6656 \\*
 &  & Biceps Femoris Tendon & 13 & 0.2150 & 0.2541 & 0.2785 & 0.3011 \\*
 &  & Popliteus Tendon & 19 & 0.5221 & 0.5709 & 0.5928 & 0.6479 \\
\addlinespace[2pt]
 & Fascia \& Muscular Compartments & Iliotibial Band & 13 & 0.1836 & 0.1945 & 0.2058 & 0.2153 \\
\addlinespace[4pt]
\textbf{A12 Sensory System} & Eye \& Ocular Adnexa & Anterior Eye Segment & 28 & 0.7485 & 0.7989 & 0.8169 & 0.8243 \\*
 &  & Posterior Eye Segment & 28 & 0.8789 & 0.8739 & 0.8763 & 0.8763 \\*
 &  & Lacrimal Gland & 28 & 0.5985 & 0.6945 & 0.6951 & 0.7035 \\
\addlinespace[2pt]
 & Ear \& Auditory-Vestibular Structures & Cochlea & 28 & 0.7173 & 0.7641 & 0.7706 & 0.7732 \\
\end{longtable}
\noindent\parbox{\textwidth}{\scriptsize\raggedright BBox denotes the initial tight 2D box on the axial slice with the largest target area. $+1$/3/5 denote 1/3/5 cumulative point refinements after the initial box. The cohort contains CT and MR only. Each row averages valid instance Dice within one Combined Target; a physical object can contribute to multiple Combined Target rows. Mean is the unweighted average of Combined Target means with valid scores.}\par
\endgroup

\bigskip
\clearpage
\begingroup
\fontsize{8}{9}\selectfont
\setlength{\tabcolsep}{3pt}
\renewcommand{\arraystretch}{1.12}
\setlength{\LTleft}{\fill}
\setlength{\LTright}{\fill}
\begin{longtable}{@{}>{\raggedright\arraybackslash}p{0.14\textwidth}>{\raggedright\arraybackslash}p{0.20\textwidth}>{\raggedright\arraybackslash}p{0.22\textwidth}rcccc@{}}
\caption{Pathological and Abnormal Targets: BBox + Point Segmentation (Dice)}\label{tab:pathology_bbox_point_target_dice} \\
\toprule
Taxonomy & Category & Combined Targets & \shortstack{Samples\\(instances)} & \multicolumn{4}{c}{\sami} \\
\cmidrule(lr){5-8}
 & & & & BBox & $+1$ & $+3$ & $+5$ \\
\midrule
\endfirsthead
\multicolumn{8}{c}{Table \thetable\ (continued)} \\
\toprule
Taxonomy & Category & Combined Targets & \shortstack{Samples\\(instances)} & \multicolumn{4}{c}{\sami} \\
\cmidrule(lr){5-8}
 & & & & BBox & $+1$ & $+3$ & $+5$ \\
\midrule
\endhead
\midrule
\multicolumn{8}{r}{Continued on next page} \\
\endfoot
\bottomrule
\endlastfoot
\textbf{Mean} &  &  &  & 0.7483 & 0.7846 & 0.8043 & 0.8132 \\
\midrule
\textbf{B01 Neoplastic Lesion} & Brain and Intracranial Tumor & Meningioma & 4 & 0.8394 & 0.9101 & 0.8987 & 0.9131 \\*
 &  & Glioma & 68 & 0.6727 & 0.7063 & 0.7334 & 0.7524 \\*
 &  & Pituitary Tumor & 10 & 0.8247 & 0.8522 & 0.8647 & 0.8689 \\*
 &  & Brain Tumor, Type Unspecified & 207 & 0.8282 & 0.8451 & 0.8542 & 0.8558 \\*
 &  & Medulloblastoma & 24 & 0.8470 & 0.8674 & 0.8653 & 0.8711 \\*
 &  & Brain Tumor with Edema Composite & 5 & 0.6897 & 0.7410 & 0.7535 & 0.7265 \\
\addlinespace[2pt]
 & Parotid Tumor & Parotid Tumor & 33 & 0.8381 & 0.8706 & 0.8670 & 0.8686 \\
\addlinespace[2pt]
 & Lung Cancer or Pulmonary Mass & Lung Cancer Lesion & 156 & 0.7978 & 0.8296 & 0.8477 & 0.8579 \\*
 &  & Primary Non-small-cell Lung Cancer & 82 & 0.5757 & 0.6196 & 0.6643 & 0.6646 \\
\addlinespace[2pt]
 & Thymic Cancer & Thymic Tumor & 70 & 0.7540 & 0.7837 & 0.8197 & 0.8380 \\
\addlinespace[2pt]
 & Breast Tumor & Breast Tumor & 257 & 0.7935 & 0.8126 & 0.8341 & 0.8442 \\*
 &  & Breast Cancer Lesion & 133 & 0.8239 & 0.8461 & 0.8573 & 0.8592 \\
\addlinespace[2pt]
 & Gastric Cancer & Gastric Cancer Lesion & 206 & 0.7046 & 0.7422 & 0.7636 & 0.7763 \\
\addlinespace[2pt]
 & Gastrointestinal Stromal Tumor & Gastrointestinal Stromal Tumor & 31 & 0.8255 & 0.8479 & 0.8653 & 0.8743 \\
\addlinespace[2pt]
 & Gastrointestinal Tumor & Colorectal Cancer Lesion & 151 & 0.6481 & 0.7004 & 0.7338 & 0.7473 \\
\addlinespace[2pt]
 & Liver Tumor & Liver Tumor & 61 & 0.7657 & 0.8228 & 0.8341 & 0.8479 \\*
 &  & Liver Cancer Lesion & 183 & 0.6963 & 0.7313 & 0.7545 & 0.7702 \\*
 &  & Hepatocellular Carcinoma Lesion & 95 & 0.6827 & 0.7923 & 0.8263 & 0.8310 \\
\addlinespace[2pt]
 & Pancreatic Tumor & Pancreatic Cancer Lesion & 30 & 0.8270 & 0.8483 & 0.8658 & 0.8708 \\*
 &  & Pancreatic Tumor & 64 & 0.8504 & 0.8761 & 0.8844 & 0.8890 \\
\addlinespace[2pt]
 & Adrenal Tumor & Adrenal Tumor & 51 & 0.8268 & 0.9070 & 0.9275 & 0.9320 \\
\addlinespace[2pt]
 & Kidney Cancer & Kidney Cancer Lesion & 39 & 0.8295 & 0.8463 & 0.8747 & 0.8747 \\
\addlinespace[2pt]
 & Kidney Tumor & Kidney Tumor & 126 & 0.8931 & 0.9094 & 0.9218 & 0.9228 \\
\addlinespace[2pt]
 & Prostate Tumor & Prostate Cancer Lesion & 14 & 0.7311 & 0.7755 & 0.7854 & 0.7847 \\
\addlinespace[2pt]
 & Cervical Cancer & Cervical Cancer Lesion & 67 & 0.7505 & 0.7737 & 0.7859 & 0.7977 \\
\addlinespace[2pt]
 & Pelvic Tumor & Pelvic Tumor, Site Unspecified & 64 & 0.7687 & 0.7834 & 0.8057 & 0.8251 \\
\addlinespace[2pt]
 & Bone Tumor & Osteosarcoma & 7 & 0.7620 & 0.8199 & 0.8330 & 0.8554 \\*
 &  & Bone Tumor, Type Unspecified & 34 & 0.7321 & 0.7694 & 0.7932 & 0.8340 \\
\addlinespace[2pt]
 & Soft-tissue Tumor & Soft-tissue Sarcoma & 41 & 0.8042 & 0.8235 & 0.8429 & 0.8485 \\*
 &  & Soft-tissue Tumor, Type Unspecified & 7 & 0.7040 & 0.7531 & 0.7817 & 0.7789 \\
\addlinespace[2pt]
 & Neurofibromatosis Type 1 (NF1) & Neurofibroma & 381 & 0.6441 & 0.6949 & 0.7213 & 0.7351 \\
\addlinespace[2pt]
 & Metastatic Tumor & Metastatic Lymph-node Tumor & 26 & 0.8189 & 0.8635 & 0.8805 & 0.8823 \\*
 &  & Liver Metastasis & 378 & 0.7309 & 0.7811 & 0.7976 & 0.8134 \\
\addlinespace[2pt]
 & Other Head and Neck Tumor & Head and Neck Primary Tumor & 150 & 0.7244 & 0.7648 & 0.7859 & 0.8014 \\*
 &  & Tongue Cancer Lesion & 17 & 0.6995 & 0.7264 & 0.7742 & 0.7742 \\
\addlinespace[2pt]
 & Other Abdominal Tumor & Abdominal Tumor, Site Unspecified & 10 & 0.7861 & 0.7836 & 0.8251 & 0.8488 \\
\addlinespace[2pt]
 & Other or Site-unspecified Tumor & Tumor Lesion, Site Unspecified & 297 & 0.7316 & 0.7618 & 0.7888 & 0.8032 \\
\addlinespace[4pt]
\textbf{B02 Infectious Lesion} & Pneumonia & COVID-19 Pulmonary Infection Lesion & 342 & 0.6834 & 0.7440 & 0.7611 & 0.7692 \\*
 &  & Pneumonia Lesion & 86 & 0.7063 & 0.7593 & 0.8043 & 0.8120 \\
\addlinespace[4pt]
\textbf{B03 Non-infectious Inflammatory \& Immune-mediated Lesion} & Central Nervous-system Demyelinating and Immune-mediated Lesion & Central Nervous System Demyelinating Lesion & 149 & 0.7420 & 0.7095 & 0.7210 & 0.7363 \\
\addlinespace[4pt]
\textbf{B04 Vascular \& Circulatory Disorder} & Intracranial Hemorrhage & Intracranial Hemorrhage & 51 & 0.6524 & 0.6583 & 0.6730 & 0.6697 \\*
 &  & Intraparenchymal Hemorrhage & 73 & 0.7457 & 0.7744 & 0.7796 & 0.7842 \\*
 &  & Intraventricular Hemorrhage & 58 & 0.6453 & 0.6931 & 0.7299 & 0.7439 \\*
 &  & Epidural Hemorrhage & 8 & 0.5720 & 0.6750 & 0.7085 & 0.7403 \\*
 &  & Subdural Hemorrhage & 36 & 0.4734 & 0.4976 & 0.4973 & 0.5028 \\*
 &  & Subarachnoid Hemorrhage & 235 & 0.6351 & 0.6802 & 0.7016 & 0.7084 \\
\addlinespace[2pt]
 & Cerebral Ischemia, Infarction and Hypoxic-ischemic Injury & Cerebral Infarction Lesion & 173 & 0.6256 & 0.6737 & 0.6985 & 0.7130 \\
\addlinespace[4pt]
\textbf{B05 Trauma \& Tissue Injury} & Fracture & Pelvic Fracture Fragment & 184 & 0.9279 & 0.9295 & 0.9556 & 0.9606 \\
\addlinespace[2pt]
 & Anterior Cruciate Ligament Injury & Anterior Cruciate Ligament Injury & 54 & 0.7102 & 0.7351 & 0.7521 & 0.7557 \\
\addlinespace[4pt]
\textbf{B07 Non-neoplastic Focal Lesion} & Renal Cyst & Renal Cyst & 84 & 0.8961 & 0.9131 & 0.9223 & 0.9258 \\
\addlinespace[4pt]
\textbf{B10 Indeterminate Imaging Abnormality} & Indeterminate Nervous-system Imaging Abnormality & Indeterminate Brain Lesion & 20 & 0.7854 & 0.8069 & 0.8205 & 0.8550 \\
\addlinespace[2pt]
 & Thyroid Nodule & Indeterminate Thyroid Lesion & 23 & 0.7831 & 0.8104 & 0.8244 & 0.8296 \\*
 &  & Thyroid Nodule & 7 & 0.7500 & 0.8012 & 0.8370 & 0.8428 \\
\addlinespace[2pt]
 & Pulmonary Nodule & Pulmonary Ground-glass Nodule & 51 & 0.8133 & 0.8450 & 0.8591 & 0.8626 \\*
 &  & Pulmonary Nodule & 350 & 0.8010 & 0.8277 & 0.8414 & 0.8470 \\*
 &  & Pulmonary Nodule and Mass Composite & 4 & 0.8343 & 0.8702 & 0.8798 & 0.8816 \\
\addlinespace[2pt]
 & Breast Nodule & Breast Nodule & 76 & 0.8424 & 0.8565 & 0.8666 & 0.8696 \\
\addlinespace[2pt]
 & Indeterminate Breast Lesion & Indeterminate Breast Lesion & 77 & 0.7067 & 0.7494 & 0.7748 & 0.7823 \\
\addlinespace[2pt]
 & Indeterminate Liver Lesion & Indeterminate Liver Lesion & 121 & 0.7687 & 0.8158 & 0.8320 & 0.8432 \\
\addlinespace[2pt]
 & Indeterminate Pancreatic Lesion & Indeterminate Pancreatic Lesion & 12 & 0.7402 & 0.7583 & 0.7493 & 0.7850 \\
\addlinespace[2pt]
 & Indeterminate Prostate Lesion & Prostate Nodule & 91 & 0.6363 & 0.7065 & 0.7370 & 0.7476 \\
\addlinespace[2pt]
 & Indeterminate Ovarian Lesion & Indeterminate Ovarian Lesion & 87 & 0.8264 & 0.8406 & 0.8626 & 0.8669 \\
\addlinespace[2pt]
 & Perirectal Lymph-node Target of Indeterminate Status & Perirectal Lymph Node of Indeterminate Status & 15 & 0.8573 & 0.8680 & 0.8589 & 0.8806 \\
\addlinespace[2pt]
 & Indeterminate Lymph-node Target & Indeterminate Lymph Node & 188 & 0.6914 & 0.7386 & 0.7529 & 0.7643 \\
\addlinespace[2pt]
 & Pathologic Fluid Collection and Abnormal Gas in Body Cavity & Pleural Effusion & 15 & 0.4129 & 0.5041 & 0.6242 & 0.6253 \\
\addlinespace[2pt]
 & Other Indeterminate Abdominopelvic Lesion & Indeterminate Kidney Lesion & 78 & 0.7840 & 0.8131 & 0.8261 & 0.8318 \\*
 &  & Indeterminate Uterine Lesion & 39 & 0.7452 & 0.8097 & 0.8280 & 0.8375 \\*
 &  & Indeterminate Bladder Lesion & 26 & 0.8247 & 0.8401 & 0.8519 & 0.8549 \\*
 &  & Adrenal Nodule & 2 & 0.7937 & 0.8525 & 0.8536 & 0.8433 \\
\end{longtable}
\noindent\parbox{\textwidth}{\scriptsize\raggedright BBox denotes the initial tight 2D box on the axial slice with the largest target area. $+1$/3/5 denote 1/3/5 cumulative point refinements after the initial box. The cohort contains CT and MR only. Each row averages valid instance Dice within one Combined Target; a physical object can contribute to multiple Combined Target rows. Mean is the unweighted average of Combined Target means with valid scores.}\par
\endgroup

\end{document}